%% file: main.tex
\documentclass{article}
\usepackage[dvipsnames]{xcolor}         
\usepackage{iclr2025_conference,times}

\usepackage[utf8]{inputenc} 
\usepackage[T1]{fontenc}    
\usepackage{hyperref}       
\usepackage{url}            
\usepackage{booktabs}       
\usepackage{amsfonts}       
\usepackage{nicefrac}       
\usepackage{microtype}      

\usepackage{xspace}
\usepackage{amsmath}
\usepackage{bbm}
\usepackage[ruled,vlined, linesnumbered, noend]{algorithm2e}
\usepackage{amssymb}
\usepackage{multirow}
\usepackage{tabularx}
\usepackage{wrapfig}
\usepackage{cleveref}
\usepackage{graphicx}
\usepackage{subcaption}
\usepackage{todonotes}
\usepackage{mathrsfs}

\usepackage{caption} 
\usepackage{array} 
\usepackage{alltt}

\input{math_commands}

\usepackage{mdframed} 
\usepackage{newfloat}

\newmdenv[
  backgroundcolor=gray!2,                  
  linecolor=gray!20,                       
  linewidth=0.5pt,                         
  roundcorner=5pt,                         
  font=\sffamily,                          
  frametitlefont=\sffamily\bfseries,       
  frametitlerule=false,                    
  frametitlealignment=\center,             
  innertopmargin=1em,                      
  innerbottommargin=1em,                   
  skipabove=1em,                           
  skipbelow=1em,                           
]{mymessagebox}

\DeclareFloatingEnvironment[
  fileext=lop,
  listname={List of Prompts},
  name=Prompt,
  placement=htp,
  within=none, 
]{prompt}

\iclrfinalcopy 
\title{On the Modeling Capabilities of Large Language Models for Sequential Decision Making}

\newcommand{\llm}{large language model\xspace}

\newcommand{\lm}{language model\xspace}

\author{%
  Martin Klissarov\footnote{Work}\\
  Mila, McGill University\\
  \And
  Devon Hjelm\\
  Apple
  \And
  Alexander Toshev\\
  Apple
  \And
  Bogdan Mazoure\\
  Apple
}

\begin{document}
\let\thefootnote\relax\footnotetext{ \textsuperscript{*} Work done during an Apple internship. Correspondence to: \texttt{martin.klissarov@mail.mcgill.ca}.}

\maketitle

\begin{abstract}
Large pretrained models are showing increasingly better performance in reasoning and planning tasks across different modalities, opening the possibility to leverage them for complex sequential decision making problems. In this paper, we investigate the capabilities of Large Language Models (LLMs) for reinforcement learning (RL) across a diversity of interactive domains. We evaluate their ability to produce decision-making policies, either directly, by generating actions, or indirectly, by first generating reward models to train an agent with RL. Our results show that, even without task-specific fine-tuning, LLMs excel at reward modeling. In particular, crafting rewards through artificial intelligence (AI) feedback yields the most generally applicable approach and can enhance performance by improving credit assignment and exploration. Finally, in environments with unfamiliar dynamics, we explore how fine-tuning LLMs with synthetic data can significantly improve their reward modeling capabilities while mitigating catastrophic forgetting, further broadening their utility in sequential decision-making tasks.
\end{abstract}

\section{Introduction}
\input{sections/1-Introduction}

\section{Using Language Models to Solve RL Tasks}
\input{sections/2-Formalization}

\section{Performance of Indirect and Direct Policy Models}
\input{sections/3-Experiments}

\section{Discussion}

\input{sections/5-Discussion}

\bibliography{main}
\bibliographystyle{iclr2025_conference}

\newpage
\appendix
\section{Appendix}
\input{sections/6-Appendix}

\section{Additional Related Works}
\input{sections/4-Related_work}

\end{document}

%% file: math_commands.tex
\usepackage{amsmath,amsfonts,bm}

\def\eqref#1{equation~\ref{#1}}

\def\1{\bm{1}}

\DeclareMathAlphabet{\mathsfit}{\encodingdefault}{\sfdefault}{m}{sl}
\SetMathAlphabet{\mathsfit}{bold}{\encodingdefault}{\sfdefault}{bx}{n}

\DeclareMathOperator*{\argmin}{arg\,min}

\newcommand{\cA}{\mathcal{A}}

\newcommand{\cD}{\mathcal{D}}

\newcommand{\cL}{\mathcal{L}}

\newcommand{\cO}{\mathcal{O}}

\newcommand{\cS}{\mathcal{S}}

\newcommand{\cX}{\mathcal{X}}

%% file: sections/1-Introduction.tex
Large Language Models (LLMs) are generative models of natural language that 
can produce accurate general and domain-specific knowledge~\citep{singhal2022large, imani2023mathprompter, manigrasso2024probing, liu2024your}, reason over long textual contexts~\citep{reid2024gemini}, and generalize zero-shot~\citep{kojima2022large}.
These capabilities suggest that LLMs might be well-suited for complex sequential decision-making problems, such as in embodied settings where an agent acts in an environment.
Recent research has begun exploring this potential, investigating how LLMs can serve as sources of intrinsic motivation~\citep{wang2024rl, klissarov2024motif}, demonstrating world modeling capabilities~\citep{lin2024learningmodelworldlanguage, liu2024world}, and for acting and/or planning directly in an environment~\citep{wang2023voyager, padalkar2023open, zhang2024can}.

However, as the predominant paradigm for training LLMs is not inherently aligned with the challenges of sequential decision-making problems, such as active exploration, it is not obvious how to best bridge their capabilities to tackle such challenges in a general manner.
We study this problem through the lens of reinforcement learning \citep[RL,][]{Sutton1998}, which formalizes how an agent interacts with an environment, receiving scalar rewards for each of its actions over a trajectory. 
We examine the capabilities of LLMs to solve RL tasks by comparing how they model policies 1) directly by generating action tokens, to 2) indirectly through a reward model derived from the LLM to be used within an RL algorithm.
We perform a comprehensive evaluation on a diverse set of domains, including MiniWob \citep{liu2018reinforcement}, NetHack \citep{kuettler2020nethack}, and Wordle \citep{Lokshtanov2022WordleIN}, and MetaWorld \citep{yu2019meta}. The environments we study present a variety of challenges, such as different action space granularities, observation modalities ranging from natural language to pixel data, and varying horizon lengths. 

We first consider the off-the-shelf capabilities of LLMs for decision-making without updating them through additional gradient updates coming from the RL task.
We find that indirectly modeling policies by first extracting knowledge from LLMs in the form of a Bradley-Terry model~\citep{Bradley1952RankAO, christiano2017deep} provides the best and most consistent performance across the environments we study. 
We empirically analyze the various benefits, and limitations, provided by this approach, showing that it improves on long-standing challenges in RL problems, such as credit assignment and exploration.

Finally, while LLMs possess knowledge useful for many decision making tasks of interest, domains with complex or unfamiliar dynamics can significantly restrict their broader utility.
We explore how fine-tuning an LLM with domain-specific data can bridge this knowledge gap and study the effect of this procedure on the LLM's previous knowledge, as measured through success on datasets like POPE~\citep{Li-hallucination-2023}, GQA~\citep{hudson2019gqa}, AI2D~\citep{kembhavi2016diagram} and MMMU~\citep{yue2024mmmu}. 
Our investigation reveals that fine-tuning for indirect policy modeling mitigates catastrophic forgetting more effectively than direct policy modeling, offering a broadly applicable strategy for leveraging LLMs across diverse sequential decision-making tasks.

%% file: sections/2-Formalization.tex
We first introduce the types of RL problems as well as formalize the methodologies for using LLMs for RL tasks used in this work.

\paragraph{Reinforcement Learning.} An RL task can be defined through a Markov Decision Process~\citep[MDP,][]{puterman2014markov}, which is composed of a state space $\cS$, an action space $\cA$, a transition function $p: \cS \times \cA \to \Delta(\cS)$ which describes the forward dynamics of the system, a reward function $r: \cS \times \cA \to \mathbb{R}$ and a discount factor $\gamma \in [0,1]$. Since it is often the case that the state is only partially observable, we also assume the environment emits an observation $o_t\sim p_\cO: \cS \to \Delta(\cO)$ from observation space $\cO$. 
A policy, or \emph{actor}, is a probability distribution $\pi: \cS \to \Delta(\cA)$ which describes the action to be taken at every step. 
The objective of a rational actor is to maximize the expected cumulative rewards over horizon $H>0$,
\begin{equation}
\label{eq:maximize}
    \max_{\pi} \mathbb{E}[\sum_{t=0}^H \gamma^t r(s_t, \pi(s_t))|s_0]=\max_\pi \mathbb{E}_{s_0}[V^\pi(s_0)],
\end{equation}

where the value function, $V^\pi(s)$, represents the expected discounted sum of rewards over the entire trajectory, re-weighted by the environment's dynamics model, $p$, and the actor's policy, $\pi$. 

\paragraph{Large Language Models.} An LLM is a generative model of discrete random variables (i.e. tokens) conditioned on a history (i.e. context). The LLM models the data distribution autoregressively:
\begin{equation}
    p(x_{t+1}|x_1, .., x_t)=\prod_{t'=1}^tp(x_{t'}|x_{<t'})\\
    =\texttt{LLM}(x_{<t},l)
\end{equation}
where $x \in \cX$ are token variables taken from a valid vocabulary. The suitability of LLMs for solving RL tasks without additional fine-tuning primarily hinges on the hypothesis that LLMs contain information -- i.e., \emph{knowledge} -- about the underlying MDP, for instance, through the policy or reward function. 
\emph{How} that information is extracted depends on the data the LLM was trained on, the ability of the practitioner to properly prompt the model and interpret its responses to solve decision-making tasks.

\subsection{Prompting}
\label{sec:prompting}
In this section, we describe the inputs, or \emph{prompts}, to the LLM used in this work which allow to change the LLM's output distribution to be useful for solving RL tasks. 
All prompts in this work use 1) task specification using natural language as input to provide information about the MDP to the LLM as context and 2) episode history in order to address issues of partial-observability in some environments~\citep[similar to the Act-only baseline prompt found in][]{yao2022react}. We additionally use the following set of techniques,

\begin{itemize}
    \item {\bf Chain of Thought}. By prompting the LLM to provide a step-by-step reasoning process for its output, rather than just the final answer, we can help surface its internal decision-making and improve the resulting performance \citep{Wei2022ChainOT}.
    \item {\bf In-Context Learning}. To enhance the LLM's ability to solve the task, example solutions (e.g., from expert policies) are provided for in-context learning \citep{Brown2020LanguageMA}, where solutions contain sequences of a combination of states, actions, and rewards.
    \item {\bf Self-Refinement}. To further refine its output, the LLM is prompted to provide recursive criticism and improvement from its generated outputs. This general strategy knows many variants, such as feedback from an environment~\citep{yao2022react}, self-critique~\citep{zelikman2022star}, or self-reflection~\citep{shinn2023reflexion}. In this work, we use Recursive Criticism and Improvement~\citep[RCI,][]{kim2024language} for its state-of-the-art performance on web agent domains and general applicability.
    In its original form, the LLM is given a task description and generates a high-level plan. This plan is used along with the task description and current state to refine an action so that it is grounded in the current observation and the action space. 
\end{itemize}

\subsection{Policy Modeling Using LLMs}
\label{sec:policy}
As shown in Equation \ref{eq:maximize}, the goal of a decision making agent is to learn a high performing policy $\pi$. This can be done either by maximizing the expected cumulative rewards and directly modeling the policy parameters \citep{Sutton1999PolicyGM,kakade2002approximately}. Equivalently, this can be done indirectly by first modeling the parameters of the value function and applying a greedy operator, such as in Q-Learning \citep{Watkins1992Qlearning}. A similar separation between direct and indirect approaches can be useful to study the capabilities of LLMs to model RL policies.

\paragraph{Direct Policy Modeling.} The most straightforward way to obtain a policy using LLMs is for the LLM to generate tokens that will be directly interpreted as actions from the environment, $a \in \mathcal{A}$~\citep{yao2022react, shinn2023reflexion, kim2024language}. To ensure the outputted actions adhere to the environment's action set, the LLM output tokens can be projected back onto $\cA$ using projection operator $\text{proj}(\cdot, \mathcal{A})$~\citep[e.g., see][for examples of projection operators]{huang2022language, kim2024language}. A variety of prompting techniques can be combined to increase the ability of the LLM to act, without task-specific fine-tuning, as a policy, which we detail in Section \ref{sec:prompting}. This direct policy method will be referred to in our experiments as \textbf{LLM Policy}.

\paragraph{Indirect Policy Modeling.} On the other hand, we can prompt the LLM to output tokens representing intermediate quantities that will then be used to learn a policy. For example, one can model the forward dynamics of the environment for planning \citep{liu2024world}, or an affordance model for action selection \citep{mullen2024towards}.
In this work, we focus on the case where these intermediate quantities will be used to generate rewards -- i.e., a \textbf{reward model} -- which will then be maximized by an off-the-shelf RL policy. 
In Section \ref{sec:reward_modeling}, we enumerate the different approaches for modeling reward functions with LLMs covered in our work.
\footnote{It is important to note that there exists many more ways in which we could indirectly model the policy. 
In Appendix \ref{sec:indirect_modeling}, we present in detail these possibilities and, in~\Cref{fig:binary_pred}, provide initial investigations that showcase their potential and limitations.}

In direct policy modeling experiments (LLM Policy), we found combining all of the prompting techniques in Section \ref{sec:prompting} to work the best, while for  indirect modeling methods through reward we relied only on chain-of-thought prompting. Additional details, such specific prompt details and ablations on these choices are presented in the Appendix \ref{sec:direct_prompts}.

\subsection{Indirectly Modeling Policies through Reward Models}
\label{sec:reward_modeling}
 We consider a diversity of methods for modeling reward functions using LLMs, with a particular attention to methods that are applicable to a diversity of environments and modalities. We study the following set,

\begin{itemize}
    \item {\bf Direct Scalar}.~  
    The LLM generates tokens that directly encode the reward (e.g., as a float or integer) given an observation (or a sequence of observations and actions). This reward is then given to the RL agent.
    \item {\bf AI Feedback} ~\citep{lee2023rlaif, klissarov2024motif}.
    Ask the LLM to express a preference $y=\{1,2,\varnothing\}$ between two observations, $o_1$ and $o_2$, for the one showing the most progress towards a certain goal, or no preference if both observations are equally good. These labels can then be collected as a dataset of observation-preference tuples $\mathcal{D}_{\text{pref}} = \{(o_1^{(i)},o_2^{(i)},y^{(i)})\}_{i=1}^M$, which are then used to train a reward function modeled as, 
\begin{equation}
    \begin{split}
        r_{\theta}=\argmin_\theta \mathbb{E}_{(o_1, o_2, y) \sim \mathcal{D}_\text{pref}}\bigg[&\mathbb{I}[y=1]\log P_\theta[o_1\succ o_2]+\mathbb{I}[y=2]P_\theta[o_2 \succ o_1]\\
        &+\frac{1}{2}\mathbb{I}[y=\varnothing]\log\big(P_\theta[o_1\succ o_2]P[o_2\succ o_1]\big)\bigg]
    \end{split}
    \label{eq:bradley_terry}
\end{equation}
where $P_\theta[o_1\succ o_2]=\frac{e^{r_\theta(o_1)}}{e^{r_\theta(o_1)}+e^{r_\theta(o_2)}}$  the probability of preferring an observation to another, referred to as the Bradley-Terry model for preference learning~\citep{Bradley1952RankAO}. The minimization of this equation is commonly done through  binary cross-entropy.
    \item {\bf Reward as Code}~\citep{Yu2023LanguageTR,Ma2023EurekaHR}. Prompt the LLM to write code that will take as input a subset of symbolic features from the environment observations and will produce a scalar output representing the reward. When symbolic features are not available, these are constructed as in \citet{Venuto2024CodeAR}.
    \item {\bf Embedding-based}~\citep{rocamonde2023vision,Du2023GuidingPI}.  Instead of querying language tokens from the LLM, we can instead, for a given input, leverage the information encoded in its latent represention, or embeddings. These embeddings are used to calculate the cosine similarity with respect to the embeddings of natural language specification of a goal or a behaviour. The resulting similarity value is given as a reward to the agent.
\end{itemize}

Additional details, such specific prompts, are presented in the Appendix \ref{sec:prompts_indirect}.

%% file: sections/3-Experiments.tex
\label{sec:indirvdir}

\begin{figure}[ht!]
    \centering
    \includegraphics[width=\linewidth]{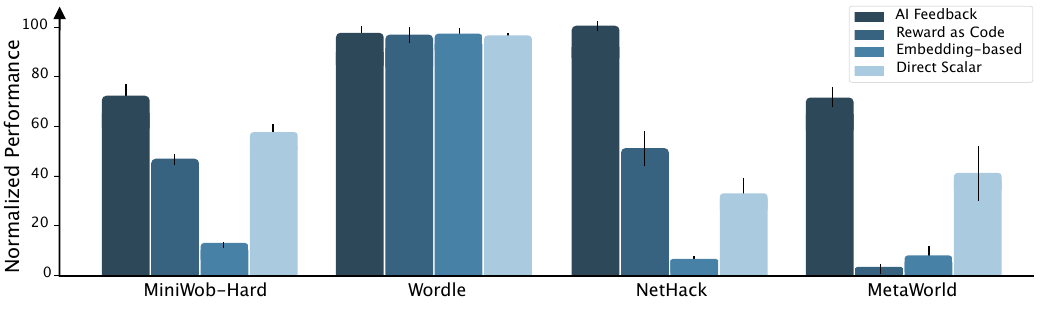}
    \caption{\textbf{AI feedback as the highest performance across different reward models derived from LLMs tested.} AI feedback, which is a preference-based method for deriving a reward model from an LLM generally outperforms other methods.}
    \label{fig:indirect_comparison}
\end{figure}

Due to fundamentally different challenges between direct and indirect policy modeling approaches, conducting a fair comparison requires care.
For example, using the LLM directly as a policy requires grounding its outputs in the action space defined by the environment~\citep{Ahn2022DoAI, huang2022language}. As the action space can vary
significantly between environments and attempting to solve this problem adds additional algorithm- or domain-specific complexities (e.g. by crafting skills, see \cite[]{Ahn2022DoAI, wang2023voyager}), we fix our experimental setting to the following

\begin{enumerate}
    \item \textbf{Atomic actions.} We only study approaches which can directly interface with the action space supported in the environment.
    In other words, the action space is at least a subspace of the space of language generated by the LLM.
    This allows for a more direct comparison across a variety of domains and study the relationship between an LLM's knowledge and the fixed action space defined by the environment. 
\item \textbf{No finetuning.} In most of the paper we assume that LLMs are used without any gradient updates, i.e. \textit{without fine-tuning} from the RL task, and evaluate their off-the-shelf capabilities.
    In~\Cref{sec:finetuning}, we perform a preliminary study on the trade-offs between fine-tuning for direct and indirect policy modeling.
\end{enumerate}


\begin{figure}[ht!]
    \centering
    \begin{subfigure}[h]{0.49\linewidth}
    \includegraphics[width=\linewidth]{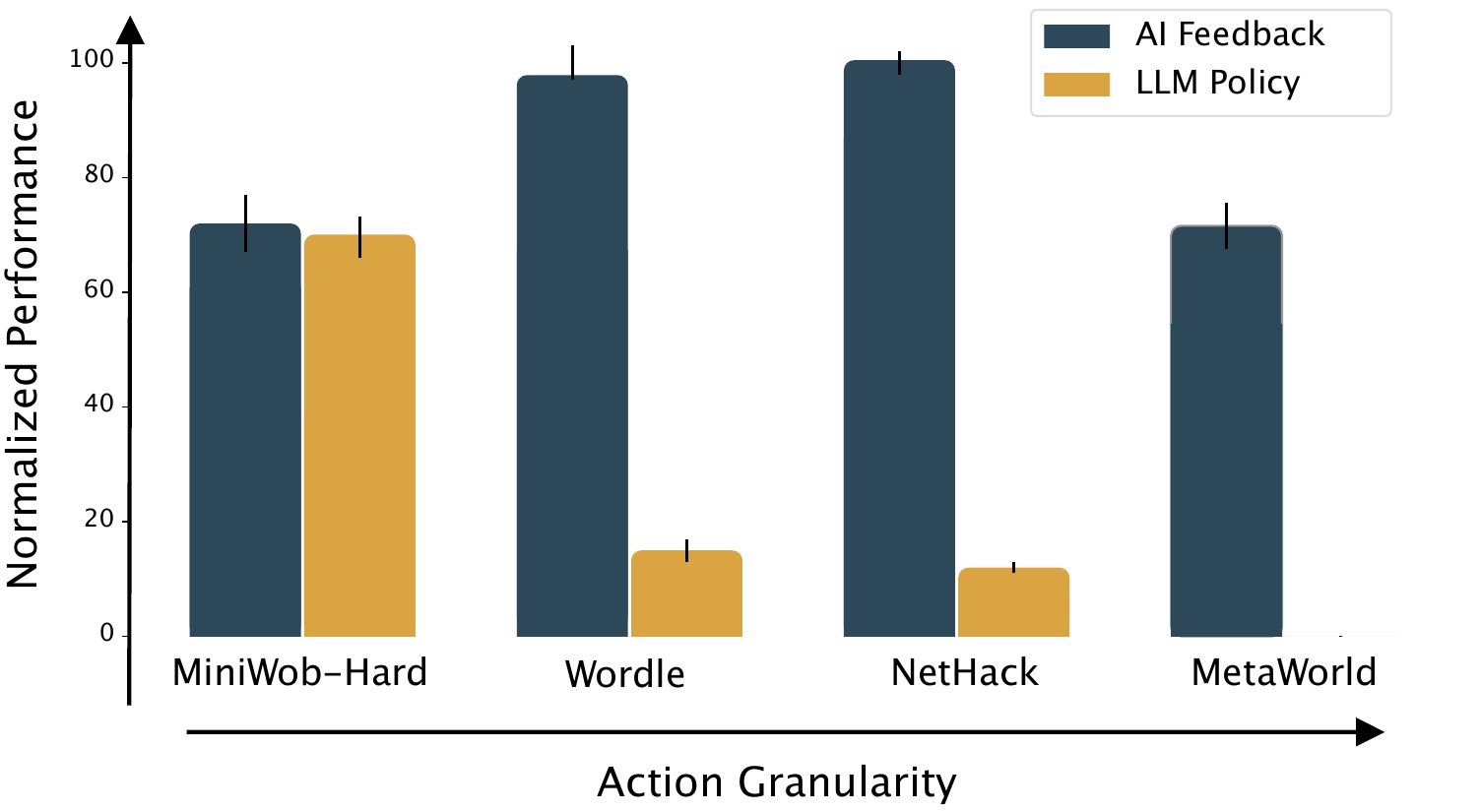}
    \caption{}
    \label{fig:actor_comparison}
    \end{subfigure}
    \begin{subfigure}[h]{0.49\linewidth}
    \includegraphics[width=\linewidth]{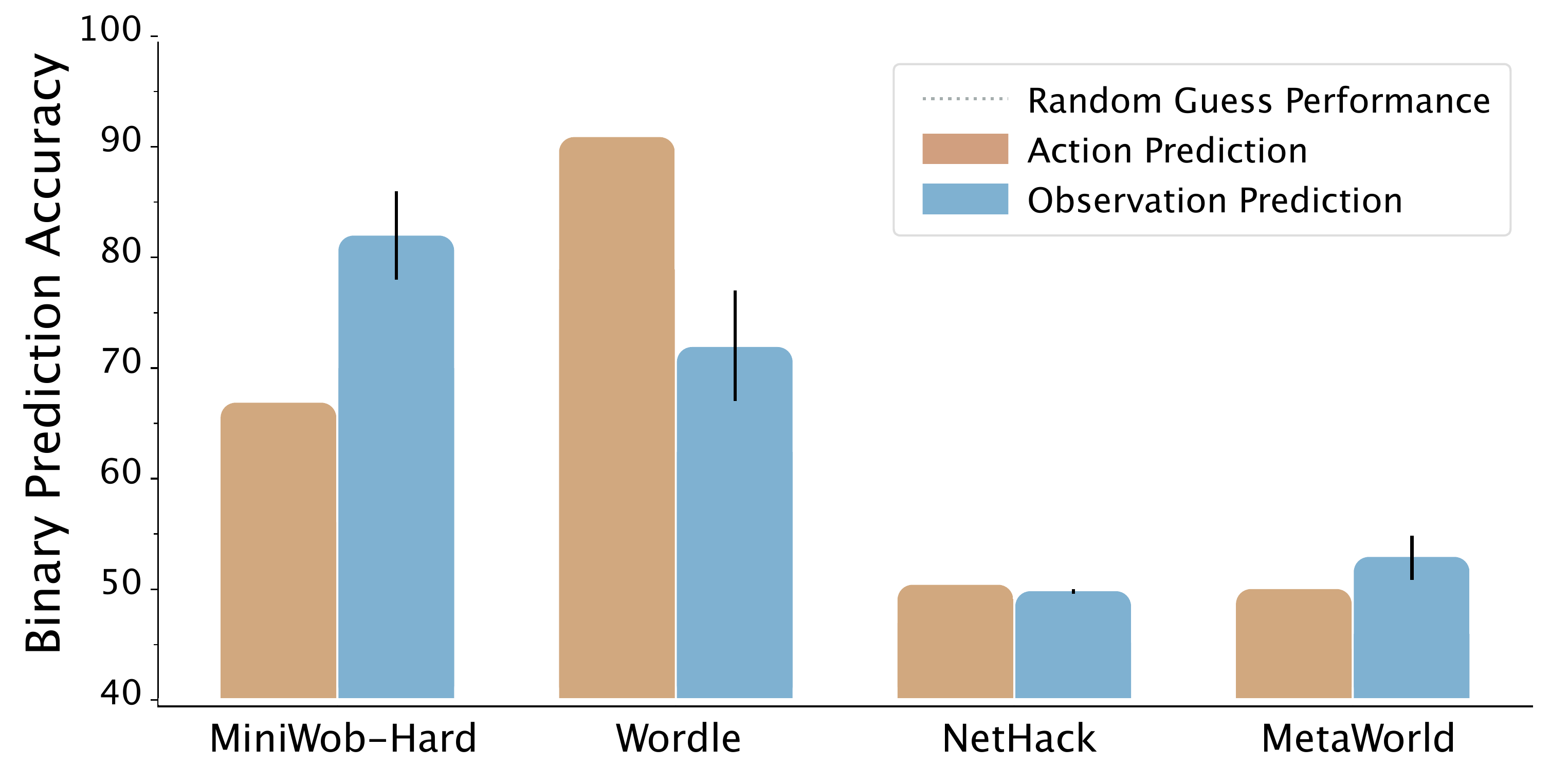}
    \caption{}
    \label{fig:binary_pred}
    \end{subfigure}
    \caption{\textbf{a) Building a reward model more-readily solves RL tasks than using an LLM as an actor.} 
    LLM-policy only performs well in domains with coarse-grained actions while LLM feedback presents strong performance across the entire range of action granularities.
    \textbf{b) LLMs have unreliable zero-shot understanding of the environment dynamics.}
    While LLMs can be used to craft useful reward models, their failure as direct policies may be explained by their poor understanding of the action space and the transition function.
    }
    \label{fig:comparison}
\end{figure}

We investigate four separate domains, where each domain aims to highlight a specific capability of LLMs: 1) MiniWob-Hard, a  subset of hard tasks from the full MiniWob suite, tests web interaction in observation/action spaces close to natural language, 2) Wordle measures reasoning and planning capabilities, 3) NetHack presents the difficulty of exploring open-ended environments under partial observability, long horizons and procedural scenarios, and 4) MetaWorld assesses the ability to control low-level, high-frequency actions in continuous space.
We provide a detailed description of each domain in Appendix \ref{sec:env_details}.  

Direct policy modeling is done by querying the closed source GPT-4o model, whereas indirect policy modeling is done through the open source models of Llama 3 \citep{dubey2024llama3herdmodels}, when environment observations consist of text, and PaliGemma \citep{Beyer2024PaliGemmaAV}, when environment observation consist of pixel images. All results are averaged over 10 seeds with error bars indicating the standard error.

\paragraph{Indirect policy modeling through rewards.} We first present a comparison of the various indirect policy modeling approaches discussed in~\Cref{sec:reward_modeling}. In these experiments, the LLM generates a reward function which will be given to a RL agent for optimization, without access to any rewards coming from the environment. When learning policies through RL we do not perform any hyperparameter search and simply borrow the existing empirical setup for each domain, as detailed in Appendix \ref{sec:env_details}.

In~\Cref{fig:indirect_comparison}, we present the performance across domains as measured by the average success rate on all domains,
except for NetHack, where performance (the in-game score) is normalized by the highest recorded value.
Results show that AI feedback is the only method that successfully crafts rewards across all environments and
modalities \footnote{In Appendix \ref{sec:extrinsic}, we verify that AI feedback yields policies with performance on par with those optimized using human-designed environment rewards.}.
On easier domains such as MiniWob-Hard, which consists of short episodes and limited scope of variations, the Direct Scalar method performs nearly as well as AI feedback.
However, the disparity between methods is much more pronounced on harder, open-ended tasks such as NetHack. 
Out of all the methods, Embedding-based leads to the lowest performance.
Finally, the effectiveness of Reward as Code appears to be highly contingent on the availability of symbolic features for code processing.
In Appendix \ref{sec:rewcode}, we further examine the assumptions—such
as access to functional knowledge of the environment—under which Reward as Code can achieve performance comparable to AI feedback

\paragraph{Direct vs indirect policy modeling.} We now compare the direct policy modeling method, LLM Policy, to the best performing indirect modeling method, AI feedback, reporting performance across the same set of domains.
Results in~\Cref{fig:actor_comparison} show that, despite the more complex prompting strategies and the use of a more capable closed source model, LLM Policy is unable to perform well in most environments, with the exception of MiniWob-Hard, where the performance is on-par with AI feedback. 

A question emerging from these results is: what factors cause this significant performance disparity between direct and indirect policy models?  One possible explanation is that LLMs, when directly queried for actions in an unfamiliar environment, may struggle to understand its dynamics (e.g., the transition function and action space). To test this hypothesis, we conduct the following experiment. We prompt the LLM to select between 1) a pair of candidate \textit{next observations} given the current observation and action (probing knowledge of $p(o_{t+1} | a_t, o_{\leq t})$), or 2) a pair of candidate \textit{actions} given the next observation and current observation (probing knowledge of $p(a_t | o_{t+1}, o_{\leq t})$). In each case, the pair contains the ground-truth and random sample. In this experiment, a $50\%$ accuracy corresponds to a random guess. 

Results presented in~\Cref{fig:binary_pred}  show that the LLM performs relatively poorly on both of these tasks, indicating limited understanding of both the action space and  the environment dynamics. This can potentially explain the limited performance of the LLM Policy approach on MiniWob-Hard, NetHack, and MetaWorld, while results on Wordle suggest that additional contributing factors are at play.

\section{Analysis of AI Feedback for RL}

Our results so far suggest that, without additional fine-tuning,
indirectly modeling policies by constructing reward functions through AI feedback is the most effective approach across the range of environments and modalities we studied.
In this section, we examine how rewards shaped by this method can assist RL agents in addressing core decision-making challenges, such as credit assignment and exploration. Through this analysis, we also emphasize the ways in which reward misspecification can unintentionally arise and severely impair performance.

\begin{figure}[ht]
\centering
\includegraphics[width=1\linewidth]{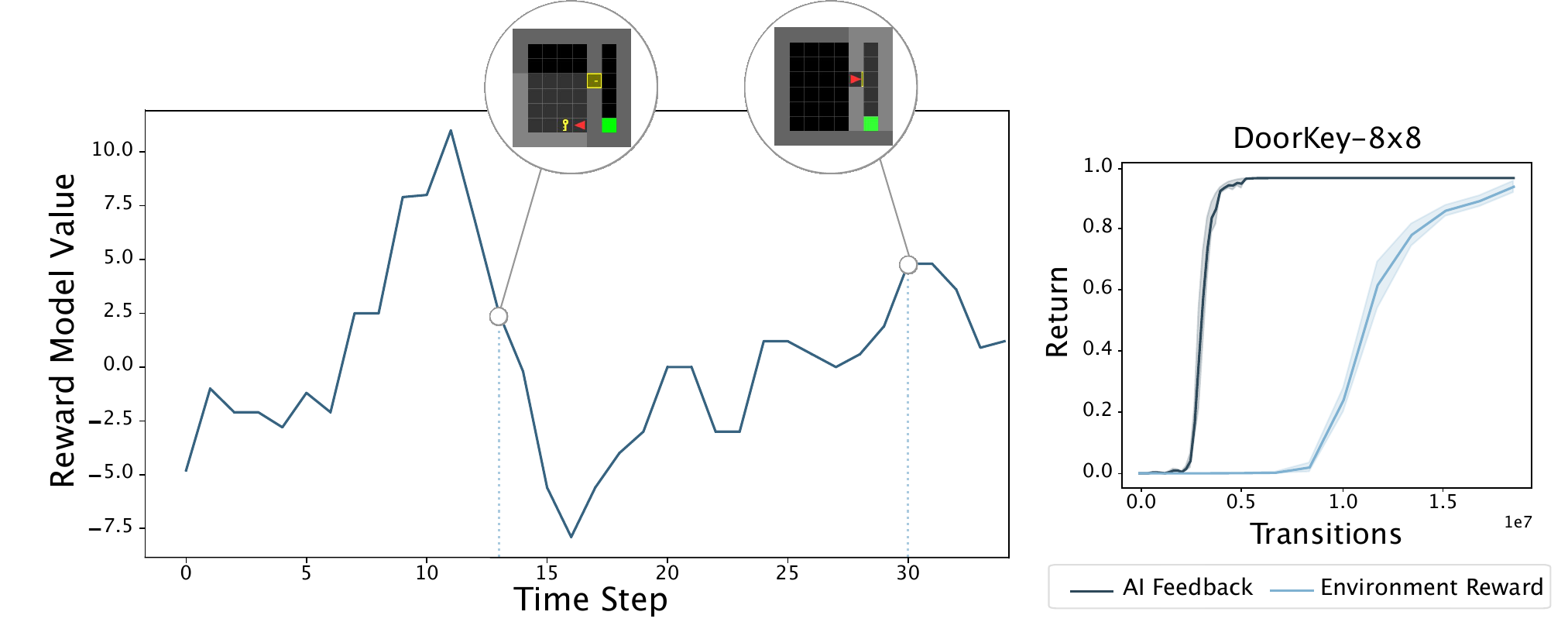}
     \caption{\textbf{Rewards learned through AI Feedback distribute rewards to key timesteps.} By doing so, the problem of credit assignment, or learning from delayed rewards, is significantly reduced. Such distribution effectively  shortens the horizon over which the RL algorithm must propagate credit through its update rule. }
    \label{fig:doorkey}
\end{figure}

\subsection{Credit Assignment}
\label{sec:credit}

AI feedback-based rewards depend on the prompt used to capture preferences. In the experiments conducted so far, these prompts were designed to elicit preferences by emphasizing states that contribute to task progress (see prompts Appendix \ref{sec:prompts_indirect}). Additionally, a key aspect of our methodology involved presenting the LLM with observations sampled randomly within trajectories. This enabled querying preference for any observation in the environment, rather limiting the focus to final states - a distinction also known as process-based and outcome-based reward models \citep{uesato2023solving,Lightman2023LetsVS}. What are the resulting characteristics of the reward model under such choices?

\paragraph{Qualitative experiment} In Figure \ref{fig:doorkey}, we present the output of the AI feedback-based reward model over each timestep of an episode within a simple grid world environment. This task includes an agent, a key, a door, and a goal \citep{MinigridMiniworld23}.
We notice that this reward model naturally captures the fact that picking up the key, as well as opening the locked door, are important steps towards the goal.
By propagating credit over such key moments in a trajectory, the LLM effectively shortens the horizon over which the RL algorithm must assign credit through temporal difference learning \citep{Sutton1998}. 
This is manifested in Figure \ref{fig:doorkey} where the agent learning through AI feedback reaches a high success rate in a fraction of the timesteps required by a similar agent learning from the environment feedback (which in this case is sparse reward of +1 for reaching the goal).

\begin{figure}[ht]
  \begin{subfigure}[t]{0.32\linewidth}
    \centering
    \includegraphics[width=\linewidth]{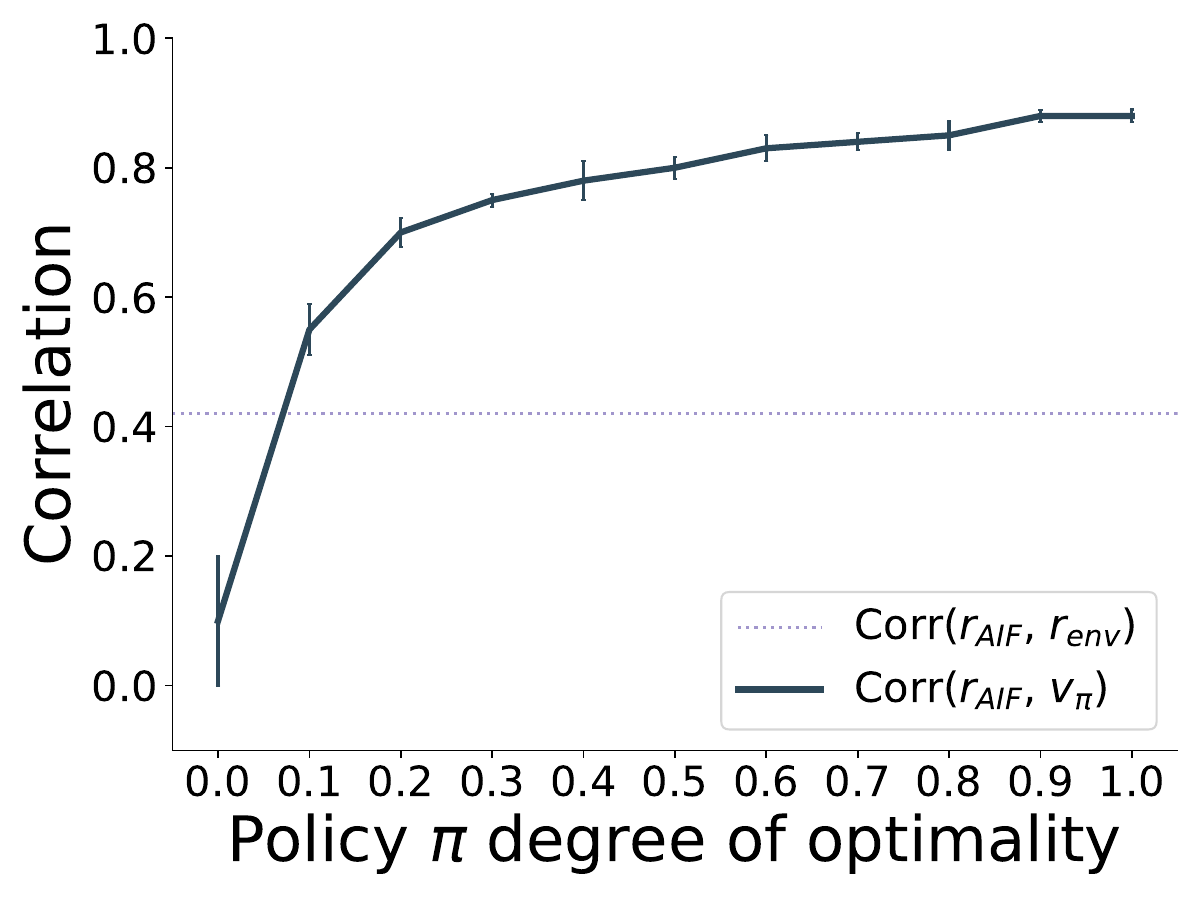}
    \caption{Wordle}
  \end{subfigure}
  \begin{subfigure}[t]{0.32\linewidth}
    \centering
    \includegraphics[width=\linewidth]{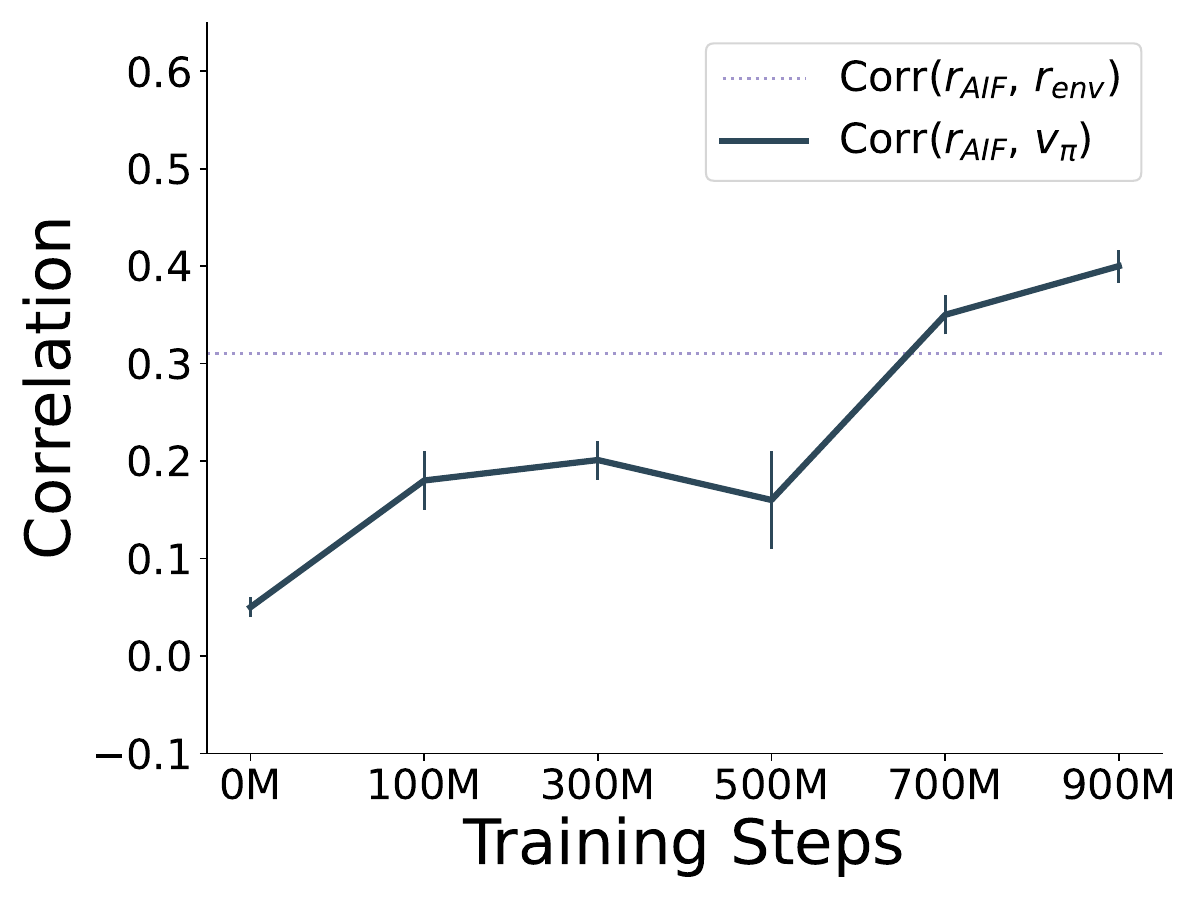}
    \caption{NetHack}
  \end{subfigure}
  \begin{subfigure}[t]{0.32\linewidth}
    \centering
    \includegraphics[width=\linewidth]{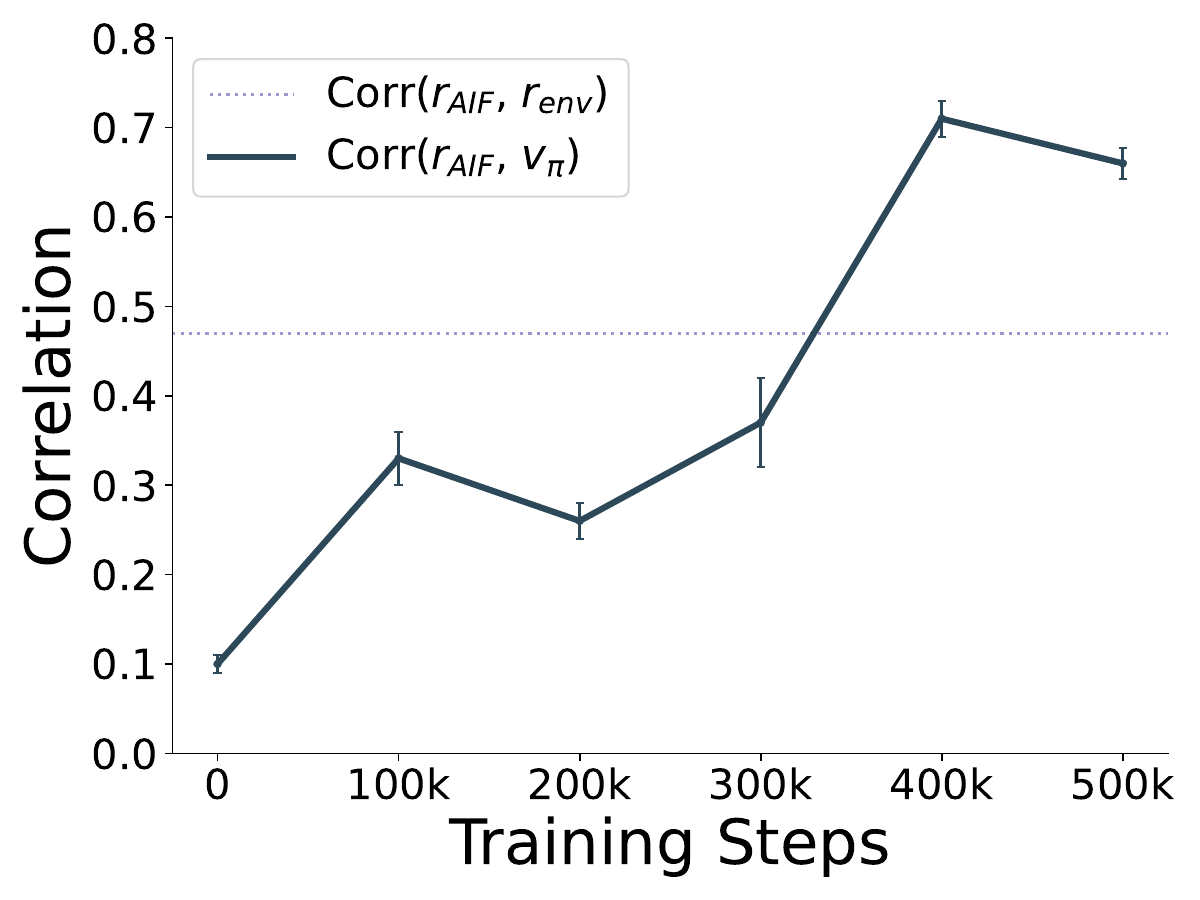}
    \caption{MetaWorld}
  \end{subfigure}
  \caption{\textbf{LLM preferences correlate with value function preferences.} The correlation between Bradley-Terry models trained from frozen LLM state preferences and value function preferences increases as the online policy improves in 3 different domains.}
  \label{fig:corr_vf}
\end{figure}

\paragraph{Quantitative experiment}
In Figure \ref{fig:corr_vf}, we present the correlation between the reward model derived from AI feedback and the value function of an RL agent across various levels of policy optimality. We observe that AI feedback generates reward functions with a stronger correlation to value functions obtained later in the training process compared to those from earlier stages. Additionally, this correlation is higher than that observed with the environment reward.
In the Wordle game, we generate, in code, a near-optimal policy and estimate its value function using Monte Carlo. We then compare it to the LLM-derived reward function find an almost perfect correlation.
These findings suggest that the reward models derived from AI feedback inherently encode aspects of high-quality value functions, which, when used as rewards for the RL agent, can substantially simplify the credit assignment process.
In Appendix \ref{sec:heuristic}, we provide additional insights from the lens of heuristic-guided reinforcement learning \citep{cheng2021heuristic}.

\subsection{Exploration}

\begin{wrapfigure}[17]{Hr}{0.4\textwidth}
    \centering
    \vspace{-1.2cm} 
    \begin{subfigure}[t]{0.95\linewidth}
        \centering
        \includegraphics[width=\linewidth]{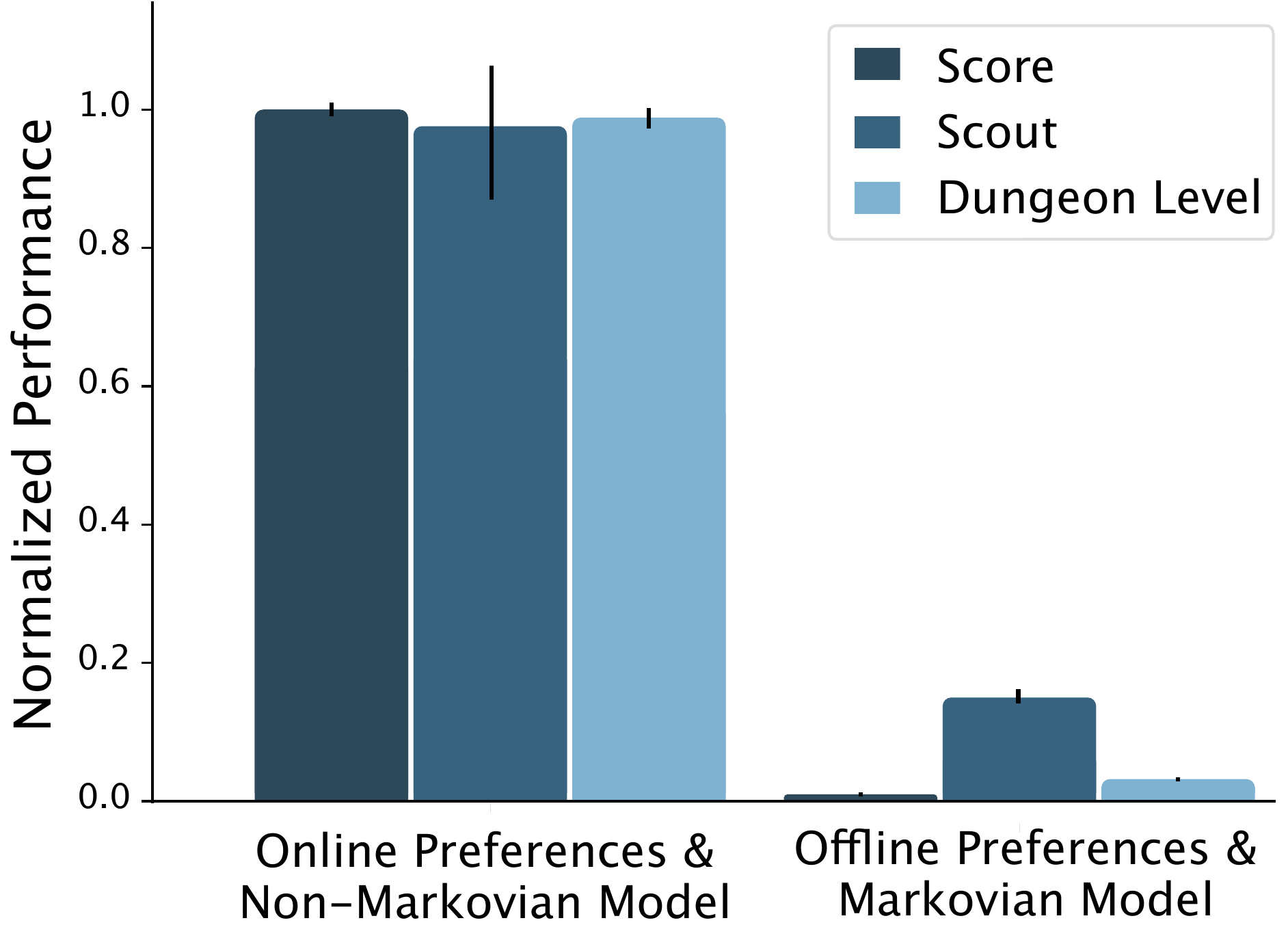}
        \label{fig:incont}
    \end{subfigure}
    \vspace{-0.3cm} 
    \captionsetup{skip=10pt}
    \caption{\textbf{By changing the prompt,
    LLMs can be steered to provide feedback that promotes exploration on NetHack.} Additionally, to avoid degenerate solutions, preferences should be elicited in an online fashion and the reward function be non-Markovian.}
    \label{fig:exploration_nethack}
\end{wrapfigure}

In the previous section, we investigated how our standard prompting strategy can ease the problem of credit assignment in downstream RL tasks. This outcome stemmed from the specific preferences we requested from the LLM, that is, promoting task progress. However, to address different RL objectives, in particular the one of exploration, we may need to elicit alternative preferences. 

Previously, \cite{klissarov2024motif} employed AI feedback to design an effective reward function for an agent operating in the open-ended environment of NetHack.
However, before applying this reward to the RL agent, the authors implemented the following transformation:
\begin{equation}
r(o_t) \propto r_{AIF}(o_t )/ N(o_t)^{\beta},
\label{eq:rint}
\end{equation}
where $r_{AIF}$ is the reward model obtained from AI feedback, $N(o_t)$ denotes the number of times a particular observation $o_t$ was seen in an episode, and $\beta$ is a positive real-valued coefficient set to $3$. 
The counting term was added to encourage exploration \citep{henaff2022exploration}, which is a key difficulty in NetHack. However, instantiating such a counting function proves difficult in many practical settings \citep{Bellemare2016UnifyingCE}. Given the flexibility of natural language, can we alleviate the need for such a term and integrate the notion of exploration in the prompt itself? 

In~\Cref{fig:exploration_nethack}, we demonstrate that this is indeed possible, leading to performance comparable when using count-based exploration by directly modifying the prompt used for preference elicitation. Specifically, when querying the LLM for preferences, we present it with a pair of sequences of observations (rather than a single observation) which provides crucial context. The prompt was also modified to steer the LLM towards avoiding low entropy sequences, i.e. sequences with repetitions (see Appendix \ref{sec:prompts_indirect}). 

Our findings reveal two potential failure modes: the offline nature of the preference elicitation method and the assumption of a Markovian reward model.
Previous research has demonstrated that online preference querying can outperform offline methods when aligning LLMs \citep{bai2022constitutional,Touvron2023Llama2O}. In our experiments, offline elicitation led to a performance collapse, likely due to frequent RL policy updates during online learning.
Additionally, assuming a Markov reward model—where the current observation fully determines the reward—can lead to an equally poor performance, as complex tasks often require historical context beyond immediate observations (see Appendix \ref{sec:nonmarkovian} for a full breakdown). 

\section{Beyond Zero-Shot Reward Modeling}
\label{sec:finetuning}

So far, we have explored the ability of LLMs to model policies, directly and indirectly, without any fine-tuning. 
However, in many cases the prior knowledge encoded in LLM might not contain the necessary information to do so successfully.  In such instances, fine-tuning becomes an effective method for incorporating task-specific knowledge into the model.

\begin{figure}[ht!]
\centering
\begin{subfigure}[t]{\linewidth}
  \includegraphics[width=\linewidth]{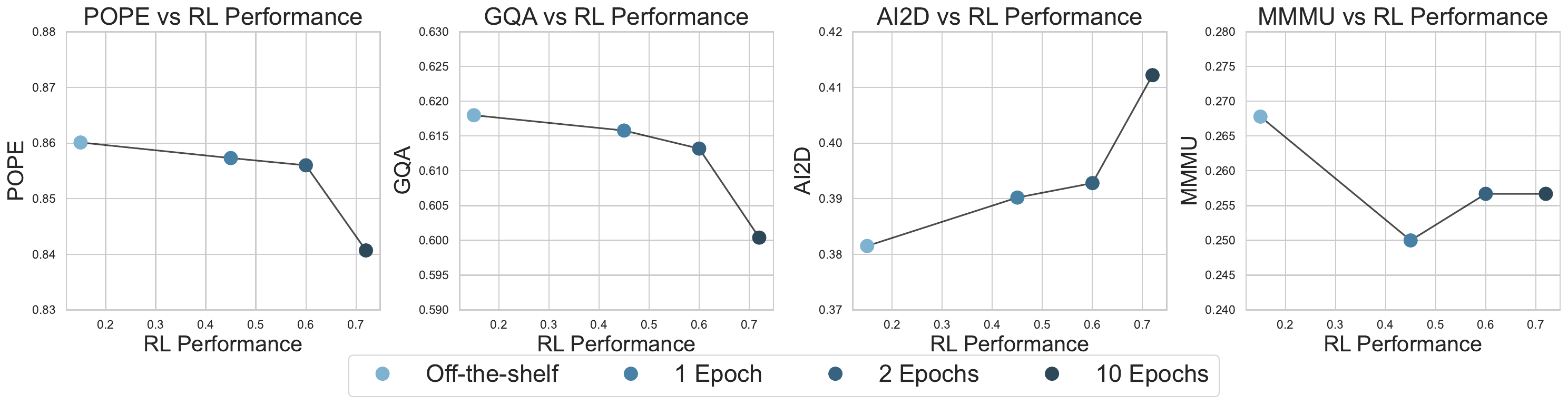} 
  \caption{Fine-tuning for AI feedback}
  \label{fig:aifeedback_finetune}
\end{subfigure}
\begin{subfigure}[t]{\linewidth}
  \includegraphics[width=\linewidth]{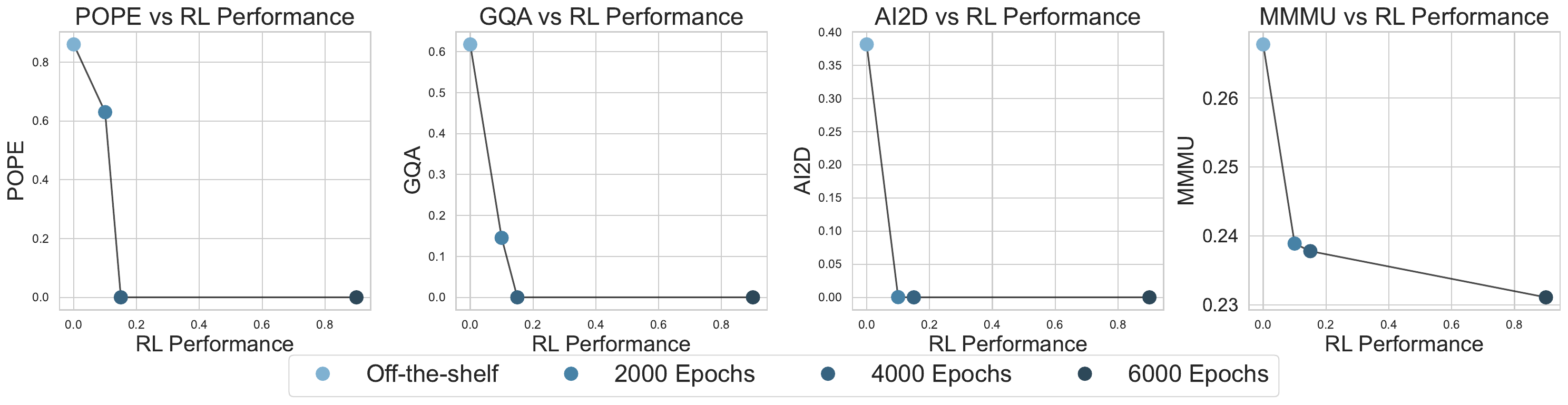} 
  \caption{Fine-tuning for direct policy modeling}
  \label{fig:actor_finetune}
\end{subfigure}
  \caption{\textbf{Fine-tuning LLMs for AI feedback better preserves their prior knowledge.} LLMs fine-tuned for AI feedback in (a) retain a higher portion of their original language reasoning knowledge than those fine-tuned for direct action selection in (b).}
  \label{fig:finetuning}
\end{figure}

We consider the \texttt{sweep-into} task from MetaWorld, where AI feedback rewards lead to a success rate of only $15\%$. When measuring the perplexity score of the PaliGemma model on captions describing the pixel observations from the task, we obtain a value of 16.03. Both of these results indicate poor understanding and the necessity to adapt the model.

We therefore fine-tune PaliGemma on image-caption pairs annotated by GPT-4o and trained the model to predict the caption for a given image. \Cref{fig:aifeedback_finetune} shows significant gains in downstream RL performance after only a few fine-tuning epochs and as few as approximately $100$ image-caption pairs. 
Moreover,~\Cref{fig:aifeedback_finetune} shows how this procedure only marginally decreases performance of the LLM on the standard multi-modal reasoning benchmarks, such as POPE~\citep{Li-hallucination-2023}, GQA~\citep{hudson2019gqa}, AI2D~\citep{kembhavi2016diagram} and MMMU~\citep{yue2024mmmu}. 
Surprisingly, performance on the AI2D benchmark~\emph{improves} as the number of task-specific fine-tuning epochs increases.

We contrast these findings with~\Cref{fig:actor_finetune}, where we fine-tune PaliGemma with behaviour cloning on expert data on the same MetaWorld task. Similarly to RT-2~\citep{brohan2023rt2visionlanguageactionmodelstransfer}, we overwrite the least frequent tokens with residual VQ-VAE codebooks~\citep{szot2024grounding}. In this case, any significant increase of RL performance comes at the cost of  catastrophically forgetting all previous knowledge. These results hint at an important trade-off: if preserving prior language reasoning knowledge is important, fine-tuning for AI feedback offers a viable approach. However, if maximizing downstream RL performance is the sole objective, directly fine-tuning for action selection can be more effective.

%% file: sections/5-Discussion.tex
In this paper, we explored two distinct approaches to leveraging LLMs for solving RL tasks: 1) directly, by modeling policies and 2) indirectly, by modeling rewards to be leveraged within a policy learning algorithm. Our results indicate that, without task-specific fine-tuning, current LLMs only show limited decision-making capabilities when directly generating actions. However, despite this limitation, LLMs are capable zero-shot reward modelers. In particular, when eliciting preferences to define rewards through the Bradley-Terry model, LLMs show strong performance across a wide range of domains presenting various challenges.

In cases where an LLM's prior knowledge is not enough to obtain useful reward functions, we also investigated fine-tuning with task-specific data to bridge this gap. 
Notably, fine-tuning to enhance reward modeling capabilities helps mitigate catastrophic forgetting, which is a crucial consideration for preserving the LLM’s general-purpose abilities
Maintaining these capabilities is essential for broad applicability to sequential decision-making tasks, including out-of-distribution tasks, and for supporting continued natural language interaction with users.

The reward modeling capabilities presented in this work offer potential solutions to challenges in RL.
First and foremost, LLM-derived reward models alleviate the need for human-designed reward functions, which are often complex and costly to develop.
Second, our empirical analysis reveals that AI-feedback based rewards produce dense functions which correlate positively with high-quality value functions.
Such reward functions can significantly reduce the difficulty of assigning credit by redistributing rewards across different steps within a trajectory.
Finally, distilling knowledge from LLMs into reward models opens new possibilities for applying RL in environments where simulators or symbolic features are unavailable—such as embodied AI agents interacting with humans.

Some notable limitations and caveats exist. For example, interacting with LLMs through natural language requires experimenting with various prompting techniques and specifications. However, this flexibility also enables the shaping of reward functions to incorporate valuable strategies \citep{Knox2013TrainingAR}, such as promoting exploration, which can further enhance the performance of RL agents.

%% file: sections/6-Appendix.tex
\label{sec:app}
\subsection{Environment Details}
\label{sec:env_details}
In our experiments, we investigate tasks from four different domains: MiniWob \citep{liu2018reinforcement}, NetHack \citep{kuettler2020nethack}, and Wordle \citep{Lokshtanov2022WordleIN}, and MetaWorld \citep{yu2019meta}. The observation space for all these environments is text, except fro MetaWorld which consists of RGB pixels.

In the MiniWob domain, we sample the subset of the five tasks on which state-of-the-art results are low. Specifically, we carry experiments on: \texttt{click-tab-2-hard}, \texttt{click-checkboxes-soft}, \texttt{count-shape}, \texttt{tic-tac-toe} and \texttt{use-autocomplete}. To learn RL policies from LLM-based rewards, we leverage the experimental setup of \citet{shaw2023pixels}. In NetHack, we use the same environment and  the same algorithmic setup as in \citet{klissarov2024motif}. In Wordle, we build on the code made available by \citet{Snell2022OfflineRF} and use their proposed subset of 200 words from the official list of the game. Finally, in MetaWorld we study the same subset of environments presented in \citep{wang2024rl} consisting of \texttt{drawer-open-v2}, \texttt{soccer-v2} and \texttt{sweep-into-v2}. Across all experiments where RL policies are learned, we use the original hyperparameter values defined in the respective experimental setups we are building upon.

\subsection{Details on Indirect Policy Modeling Through LLM-based Rewards}
\label{sec:prompts_indirect}

We use the following prompt templates to query the agent for AI feedback, Scalar Reward and Reward as Code across various environments. For the Embedding-based approach, we use calculate the cosine similarity between the representation, provided by a BERT \citep{Devlin2019BERTPO} sentence encoder (specifically the same \texttt{paraphrase-MiniLM-L3-v2} model) when environments are text-based, and otherwise we use the CLIP encoder \citep{Radford2021LearningTV}. The similarity is measured between the current observation and the same goal description contained in the each of the following prompts given for the other baselines.

\begin{prompt}[h!]
\centering
\begin{mymessagebox}[frametitle=MiniWob Prompt For Reward Modeling with AI feedback]
\small\fontfamily{pcr}\selectfont
I will present you with two HTML descriptions from a web interaction environment.\\
\newline
\{task\_description\}
\newline
Write an analysis describing the semantics of each description strictly using information from the descriptions.\\
Provide a comparative analysis based on first principles.\\
Finally, express a preference based on which description is the most likely to make some progress towards the goal,
writing either ("best\_description": 1), ("best\_description": 2).\\
You could also say ("best\_description": None).\\
\newline
html\_description\_1: \{description\_1\} \\

html\_description\_2: \{description\_2\} \\

    \end{mymessagebox}
    \caption{}
    \label{prm:miniwob_aif}
\end{prompt}

\begin{prompt}[h!]
\centering
\begin{mymessagebox}[frametitle=Wordle Prompt For Reward Modeling with AI feedback]
\small\fontfamily{pcr}\selectfont
I will present you with two short gameplay descriptions of Wordle.\\
\newline
First, tell me about your knowledge of Wordle.\\
\newline
Mention the goal of Wordle. Use the following information for Wordle states:
black means that the provided letter is not present anywhere in the hidden word.
yellow means that the provided letter is present somewhere in the hidden word, but not at the correct position.
green means that the provided letter is present in the hidden word exactly at the correct position.\\
\newline
Then, write an analysis describing the semantics of each description strictly using information from the descriptions (which may be empty) and your knowledge of Wordle.\\
Provide a comparative analysis based on first principles.\\
Finally, express a preference based on which description is the most likely to make some progress towards the goal,
writing either ("best\_description": 1), ("best\_description": 2).\\
You could also say ("best\_description": None).\\
\newline
description\_1: \{description\_1\} \\

description\_2: \{description\_2\} \\

    \end{mymessagebox}
    \caption{}
    \label{prm:rlaif_wordle}
\end{prompt}

\begin{prompt}[h!]
\centering
\begin{mymessagebox}[frametitle=NetHack Prompt For Reward Modeling with AI feedback]
\small\fontfamily{pcr}\selectfont
I will present you with two short gameplay descriptions of Nethack.\\
\newline
First, tell me about your knowledge of NetHack.\\
\newline
Mention the goal of NetHack. Prefer agents that maximize the score in the game, for instance by killing monsters, collecting gold or going down the stairs in the dungeon.\\
\newline
Then, write an analysis describing the semantics of each description strictly using information from the descriptions (which may be empty) and your knowledge of NetHack.\\
Provide a comparative analysis based on first principles.\\
Finally, express a preference based on which description is the most likely to make some progress towards the goal,
writing either ("best\_description": 1), ("best\_description": 2).\\
You could also say ("best\_description": None).\\
\newline
description\_1: \{description\_1\} \\

description\_2: \{description\_2\} \\

    \end{mymessagebox}
    \caption{}
    \label{prm:rlaif_nle}
\end{prompt}

\begin{prompt}[h!]
\centering
\begin{mymessagebox}[frametitle=NetHack Prompt For Online Reward Modeling with AI feedback]
\small\fontfamily{pcr}\selectfont
I will present you with two short gameplay descriptions of Nethack.\\
\newline
First, tell me about your knowledge of NetHack.\\
\newline
Mention the goal of NetHack. Prefer agents that maximize the score in the game, for instance by killing monsters, collecting gold or going down the stairs in the dungeon.\\
\newline
Then, write an analysis describing the semantics of each description strictly using information from the descriptions (which may be empty) and your knowledge of NetHack.\\
Provide a comparative analysis based on first principles.\\
Finally, express a preference based on which description is the most likely to make some progress towards the goal,
writing either ("best\_description": 1), ("best\_description": 2).\\
You could also say ("best\_description": None).\\
\newline
description\_1: \{description\_1\} \\

description\_2: \{description\_2\} \\

    \end{mymessagebox}
    \caption{}
    \label{prm:repetitions}
\end{prompt}

\begin{prompt}[h!]
\centering
\begin{mymessagebox}[frametitle=MetaWorld Prompt For Reward Modeling with AI feedback]
\small\fontfamily{pcr}\selectfont
Does the image satisfy \{current\_task\}?\\
image\_1: \{image\_1\} \\
\textcolor{blue}{\{llm\_response\}}\\
\\
Does the image satisfy \{current\_task\}?\\
image\_2: \{image\_2\} \\
\textcolor{blue}{\{llm\_response\}}
    \end{mymessagebox}
    \caption{}
    \label{prm:metaworld_aif}
\end{prompt}

\begin{prompt}[h!]
\centering
\begin{mymessagebox}[frametitle=MiniWob Prompt For Reward Modeling with Scalar Reward]
\small\fontfamily{pcr}\selectfont
I will present you with an HTML descriptions from a web interaction environment.\\
\newline
\{task\_description\}\\
\newline
Write an analysis describing the semantics of the description strictly using information from the description.\\

Finally, output a scalar value between $0$ and $5$, with higher values correlation with  progress towards the goal.\\
\\
html\_description: \{description\} \\

    \end{mymessagebox}
    \caption{}
    \label{prm:miniwob_sr}
\end{prompt}

\begin{prompt}[h!]
\centering
\begin{mymessagebox}[frametitle=Wordle Prompt For Reward Modeling with Scalar Reward]
\small\fontfamily{pcr}\selectfont
I will present you with a gameplay description of Wordle.\\
\newline
First, tell me about your knowledge of Wordle.\\
\newline
Mention the goal of Wordle. Use the following information for Wordle states:
black means that the provided letter is not present anywhere in the hidden word.
yellow means that the provided letter is present somewhere in the hidden word, but not at the correct position.
green means that the provided letter is present in the hidden word exactly at the correct position.\\
\newline
Write an analysis describing the semantics of the description strictly using information from the description.\\
Finally, output a scalar value between $0$ and $5$, with higher values correlation with  progress towards the goal.\\
description: \{description\} \\

    \end{mymessagebox}
    \caption{}
    \label{prm:wordle_sr}
\end{prompt}

\begin{prompt}[h!]
\centering
\begin{mymessagebox}[frametitle=NetHack Prompt For Reward Modeling with Scalar Reward]
\small\fontfamily{pcr}\selectfont
I will present you with a gameplay description of Nethack.\\
\newline
First, tell me about your knowledge of NetHack.\\
\newline
Mention the goal of NetHack. Prefer agents that maximize the score in the game, for instance by killing monsters, collecting gold or going down the stairs in the dungeon.\\
\newline
Write an analysis describing the semantics of the description strictly using information from the description.\\
Finally, output a scalar value between $0$ and $5$, with higher values correlation with  progress towards the goal.\\
description: \{description\} \\

    \end{mymessagebox}
    \caption{}
    \label{prm:sr_nle}
\end{prompt}

\begin{prompt}[h!]
\centering
\begin{mymessagebox}[frametitle=MetaWorld Prompt For Reward Modeling with Scalar Reward]
\small\fontfamily{pcr}\selectfont
From 0 to 5, how much does the image achieve\{current\_task\}?\\
image: \{image\}
    \end{mymessagebox}
    \caption{}
    \label{prm:metaworld_sr}
\end{prompt}

\begin{prompt}[h!]
\centering
\begin{mymessagebox}[frametitle=MiniWob Prompt For Reward Modeling with Reward as Code]
\small\fontfamily{pcr}\selectfont
I will present you with HTML descriptions from a web interaction environment.\\
\newline
\{task\_description\}\\
\newline
Write an analysis describing the semantics of  the descriptions strictly using information from the descriptions.\\

Finally, write a code that, when executed, will help make progress towards the goal.\\
\\
html\_descriptions: \{descriptions\} \\

    \end{mymessagebox}
    \caption{}
    \label{prm:miniwob_code}
\end{prompt}

\begin{prompt}[h!]
\centering
\begin{mymessagebox}[frametitle=Wordle Prompt For Reward Modeling with Reward as Code]
\small\fontfamily{pcr}\selectfont
I will present you with gameplay descriptions of Wordle.\\
\newline
First, tell me about your knowledge of Wordle.\\
\newline
Mention the goal of Wordle. Use the following information for Wordle states:
black means that the provided letter is not present anywhere in the hidden word.
yellow means that the provided letter is present somewhere in the hidden word, but not at the correct position.
green means that the provided letter is present in the hidden word exactly at the correct position.\\
\newline
Write an analysis describing the semantics of the descriptions strictly using information from the description.\\
Finally, write a code that, when executed, will help make progress towards the goal.\\
descriptions: \{descriptions\} \\

    \end{mymessagebox}
    \caption{}
    \label{prm:rlaif_wordle}
\end{prompt}

\begin{prompt}[h!]
\centering
\begin{mymessagebox}[frametitle=NetHack Prompt For Reward Modeling with Reward as Code]
\small\fontfamily{pcr}\selectfont
I will present you with gameplay descriptions of Nethack.\\
\newline
First, tell me about your knowledge of NetHack.\\
\newline
Mention the goal of NetHack. Prefer agents that maximize the score in the game, for instance by killing monsters, collecting gold or going down the stairs in the dungeon.\\
\newline
Write an analysis describing the semantics of the descriptions strictly using information from the descriptions.\\
Finally, write a code that, when executed, will help make progress towards the goal.\\
descriptions: \{descriptions\} \\

    \end{mymessagebox}
    \caption{}
    \label{prm:code_nle}
\end{prompt}

\subsection{Details on Direct Policy Modeling}
\label{sec:direct_prompts}
We present the exact prompts used to query GPT-4o for each of the domains we have considered. These are presented through Prompt \ref{prm:miniwob}, \ref{prm:nethack}, \ref{prm:wordle} and \ref{prm:metaworld}.

\begin{figure}
    \centering
    \includegraphics[width=0.5\linewidth]{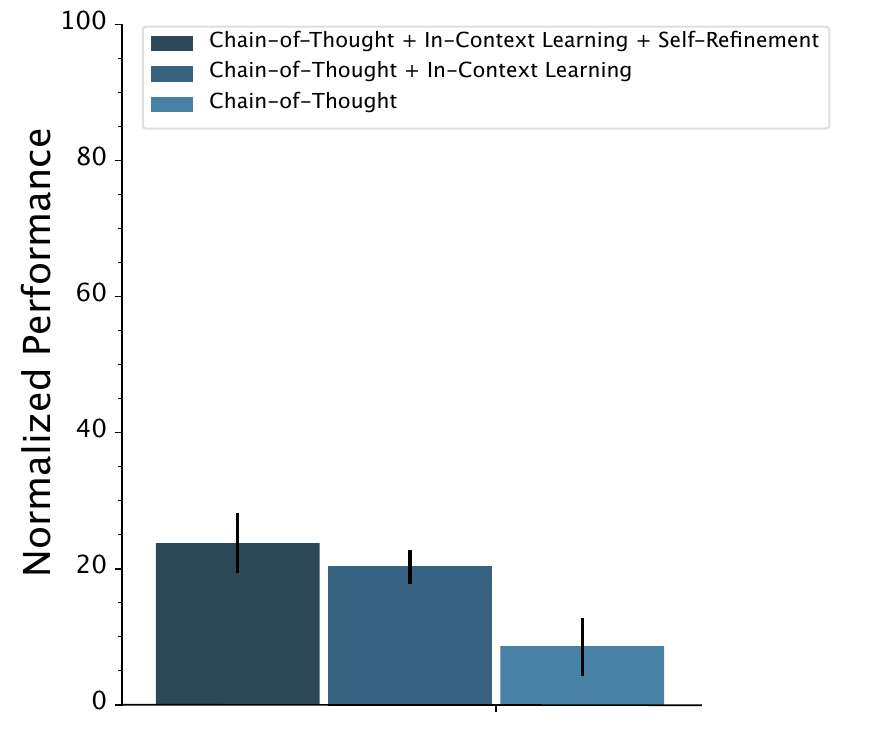}
    \caption{Ablation on the set of prompting techniques used for direct policy modeling. The reported performance is averaged over all domains and tasks.}
    \label{fig:direct_ablate}
\end{figure}
Additionally, in Figure \ref{fig:direct_ablate}, we ablate the prompting techniques used in our direct policy modeling approach. Results show that a combination of all prompting techniques presened in Section \ref{sec:prompting} works best.

\begin{prompt}[h!]
\centering
\begin{mymessagebox}[frametitle=MiniWob-Hard Prompt For Direct Policy Modeling]
\small\fontfamily{pcr}\selectfont
We have an autonomous computer control agent that can perform atomic instructions specified by natural language to control computers. There are two types of instructions it can execute. \\
\\
First, given the instruction that matches the regular expression, \text{"\textasciicircum type.\{1,\}\$"}
\\
Second, given the instruction that matches the regular expression, \text{"\textasciicircum clickxpath\textbackslash s.\{1,\}\$"}
it can click an HTML element with an xpath that is visible on the webpage. The target of this instruction should be a valid xpath. \\
\\
Below is the HTML code of the webpage where the agent should solve a task. \\
\\
\{html\_observation\} \\
\\
Examples: \\
task: \{example\_task\}\\
plan: \{example\_plan\}\\
\\
Current task: Enter an item that starts with \"Anti\" and ends with \"da\". \\
Think step-by-step before answering, what is the current plan? \{\textcolor{blue}{llm\_plan}\} \\
\\
=============== \\
Repeat N times: \\
\\
Find problems with this plan for the given task compared to the example plans. \\
\\
\{\textcolor{blue}{llm\_criticism}\} \\
\\
Based on this, what is the plan for the agent to complete the task? \\
\\
Below is the HTML code of the webpage where the agent should solve a task. \\
  \{html\_observation\} \\
\\
Current task: Enter an item that starts with \"Anti\" and ends with \"da\". \\
Think step-by-step before answering, what is the current plan? \{\textcolor{blue}{llm\_plan}\} \\
=============== \\
    \end{mymessagebox}
    \caption{
    }
    \label{prm:miniwob}
\end{prompt}

\begin{prompt}[h!]
\centering
\begin{mymessagebox}[frametitle=Wordle Prompt for Direct Policy Modeling]
\small\fontfamily{pcr}\selectfont
Let's play a game of Wordle. You will have to guess the words and I will give you the colors.\\
\\
Use the following information for Wordle colors:\\
black means that the provided letter is not present anywhere in the hidden word.\\
yellow means that the provided letter is present somewhere in the hidden word, but not at the correct position.\\
green means that the provided letter is present in the hidden word exactly at the correct position.\\
\\
You can choose among this list of words: \{list\_of\_words\}\\
\\
Here are examples trajectories, containing past observations and actions, together with an appropriate action.\\
\\
Example 1:\\
Trajectory: \{example\_trajectory\}\\
Action: \{example\_action\}\\
\\
Example 2:\\
Trajectory: \{example\_trajectory\}\\
Action: \{example\_action\}\\
\\
Current trajectory: \{trajectory\_so\_far\}\\
Think step-by-step before answering, what should be the current action? \{\textcolor{blue}{llm\_action}\}\\
\\
==============\\
Repeat N times:\\
\\
Find problems with this action for the given task compared to the example actions.\\
\\
\{\textcolor{blue}{llm\_criticism}\}\\
\\
Based on this, what is the action for the agent to make progress on the task?\\
\\
Current trajectory: \{trajectory\_so\_far\}\\
Think step-by-step before answering, what should be the current action? \{\textcolor{blue}{llm\_action}\}\\
==============\\

    \end{mymessagebox}
    \caption{
    }
    \label{prm:wordle}
\end{prompt}

\begin{prompt}[h!]
\centering
\begin{mymessagebox}[frametitle=NetHack Prompt for Direct Policy Modeling]
\small\fontfamily{pcr}\selectfont
Let's play the game of NetHack. 
\\
\\First, tell me about your knowledge of NetHack. Mention the goal of NetHack. 
\\
\\Prefer maximizing the score in the game, for instance by killing monsters, collecting gold or going down the stairs in the dungeon.
\\
\\Here are examples sub-trajectories, containing past observations and actions, together with an appropriate action.
\\
\\Example 1: 
\\sub-Trajectory: \{example\_sub-trajectory\}
\\Action: \{example\_action\}
\\
\\Example 2:
\\sub-Trajectory: \{example\_sub-trajectory\}
\\Action: \{example\_action\}
\\
\\Current sub-trajectory: \{sub-trajectory\_so\_far\}
\\Think step-by-step before answering, what should be the current action? \{\textcolor{blue}{llm\_action}\}
\\
\\==============
\\Repeat N times:
\\
\\Find problems with this action for the given task compared to the example actions.
\\
\\\{\textcolor{blue}{llm\_criticism}\}
\\
\\Based on this, what is the action for the agent to make progress on the task?
\\
\\Here is the current sub-trajectory, containing past observations and actions: \{sub-trajectory\_so\_far\}
\\Think step-by-step before answering, what should be the current action? \{\textcolor{blue}{llm\_action}\}
\\==============
    \end{mymessagebox}
    \caption{
    }
    \label{prm:nethack}
\end{prompt}

\begin{prompt}[h!]
\centering
\begin{mymessagebox}[frametitle=MetaWorld Prompt for Direct Policy Modeling]
\small\fontfamily{pcr}\selectfont
You are controlling a robot for the following task: 
\\
\\\{meta\_world\_task\} 
\\
\\Here are examples sub-trajectories, containing past observations and actions, together with an appropriate action.
\\
\\Example 1: 
\\sub-Trajectory: \{example\_sub-trajectory\}
\\Action: \{example\_action\}
\\
\\Example 2:
\\sub-Trajectory: \{example\_sub-trajectory\}
\\Action: \{example\_action\}
\\
\\Current sub-trajectory: \{sub-trajectory\_so\_far\}
\\Think step-by-step before answering, what should be the current action? \{\textcolor{blue}{llm\_action}\}
\\
\\==============
\\Repeat N times:
\\
\\Find problems with this action for the given task compared to the example actions.
\\
\\\{\textcolor{blue}{llm\_criticism}\}
\\
\\Based on this, what is the action for the agent to make progress on the task?
\\
\\Here is the current sub-trajectory, containing past observations and actions: \{sub-trajectory\_so\_far\}
\\Think step-by-step before answering, what should be the current action? \{\textcolor{blue}{llm\_action}\}
\\==============

    \end{mymessagebox}
    \caption{
    }
    \label{prm:metaworld}
\end{prompt}

\subsection{Additional Indirect Policy Modeling Methods}
\label{sec:indirect_modeling}



There are a number of other prompting methods for extracting information or \emph{knowledge} from an LLM that may be relevant to solving RL tasks.
\begin{itemize}
    \item {\bf Direct State Generation}. The model generates tokens that will represent next states (or other-future-time states). This is similar to world modeling. The next state prediction can be conditioned on an action, or marginalized over a policy distribution.
    \item {\bf  Action Preference}. Ask the LLM to select, among two choices, the most likely action given previous and future observations.
    \item {\bf  State Preference}. Ask the LLM to select, among two choices, the most likely next state or observation conditioned on prior history and/or actions.
\end{itemize}

Many of the above could theoretically be used to construct a policy, yet a full implementation is out of scope from this paper due to the lack of available code-bases to build upon and we do not seek to build new algorithms from scratch. However, in Figure \ref{fig:binary_pred} we perform investigations into the capabilities of LLMs to perform  Action Preference and  State Preference. The results show that current LLMs struggle to achieve strong performance on any of these tasks. Additionally, in Table \ref{tab:next_obs_pred}, we report the accuracy with which LLMs directly predicts the next observation (Direct State Generation), providing a probe into their direct world modeling capabilities.
Results show limited performance, except on MiniWob-Hard tasks, which are fully observable and encode deterministic transitions. 

\begin{table}
\begin{center}
\vspace{-0.5cm}
\begin{tabularx}{0.35\textwidth}{>{\raggedright\arraybackslash}X r}
\toprule
 & Accuracy \\
\midrule
MiniWob-Hard & $65 \pm 11.4 \%$ \\
Wordle & $28 \pm 8.3\%$ \\
NetHack & $0.0 \pm 0.0 \%$ \\
MetaWorld & N/A \\
\bottomrule
\end{tabularx}
\end{center}
\caption{{\bf LLMs struggle to predict the next observation.} We show the decreasing accuracy of the LLM to predict the next observation with increasing task complexity. LLMs are unable to generate pixel observations, which are used in MetaWorld.}
\label{tab:next_obs_pred}
\end{table}



\subsection{Ablating Reward as Code}
\label{sec:rewcode}

\begin{table}[ht!]
\begin{center}
\begin{tabularx}{0.65\textwidth}{>{\raggedright\arraybackslash}X r}
\toprule
\multicolumn{2}{c}{Reward as Code - RGB Observations} \\
\midrule
GPT-4o &  \\
\quad \quad w/o expert demonstration & $0\% \pm 1\%$ \\
\quad \quad with expert demonstration & $79\% \pm 7\%$ \\
\midrule
\multicolumn{2}{c}{Reward as Code - Proprioceptive Observations} \\
\midrule
Llama 3 70B &  \\
\quad \quad w/o background functional knowledge & $0\% \pm 1\%$ \\
\quad \quad with background functional knowledge & $10\% \pm 3\%$ \\
 GPT-4o &  \\
\quad \quad w/o background functional knowledge & $5\% \pm 3\%$ \\
\quad \quad with background functional knowledge & $76\% \pm 6\%$ \\
\midrule
\multicolumn{2}{c}{AI Feedback - RGB Observations} \\
\midrule
\quad PaliGemma & $72\% \pm 8\%$ \\
\bottomrule
\end{tabularx}
\end{center}
\caption{\textbf{AI Feedback performs on par with Reward as Code, without proprioceptive observations or expert demonstrations.}  To match AI Feedback performance on Metaworld, Reward as Code requires GPT-4o level knowledge, augmented with either in-context expert demonstrations or proprioceptive observations.}
\label{tab:rewcode}
\end{table}

In Table \ref{tab:rewcode}, we ablate the performance of the Reward as Code baseline across LLMs, observation spaces and additional assumptions. For pixel observations, we follow the methodology laid out in \citep{Venuto2024CodeAR}, whereas for proprioceptive observations we follow the one from \citep{Yu2023LanguageTR}. Both methods heavily depend on access to a state-of-the-art, closed-source model to achieve performance comparable to that of AI Feedback, which uses the smaller, open-source model of Paligemma \citep{Beyer2024PaliGemmaAV}. Additionally, each method requires expert demonstrations or specialized domain knowledge to guide the reward design process. While these assumptions may be viable in certain situations, such as in a controlled simulation environment, they can present significant practical challenges in more general contexts. In contrast, AI Feedback operates by simply comparing observations and reasoning using a chain-of-thought approach.

\subsection{Learning from Environment Rewards}
\label{sec:extrinsic}
\begin{figure}
    \centering
    \includegraphics[width=0.8\linewidth]{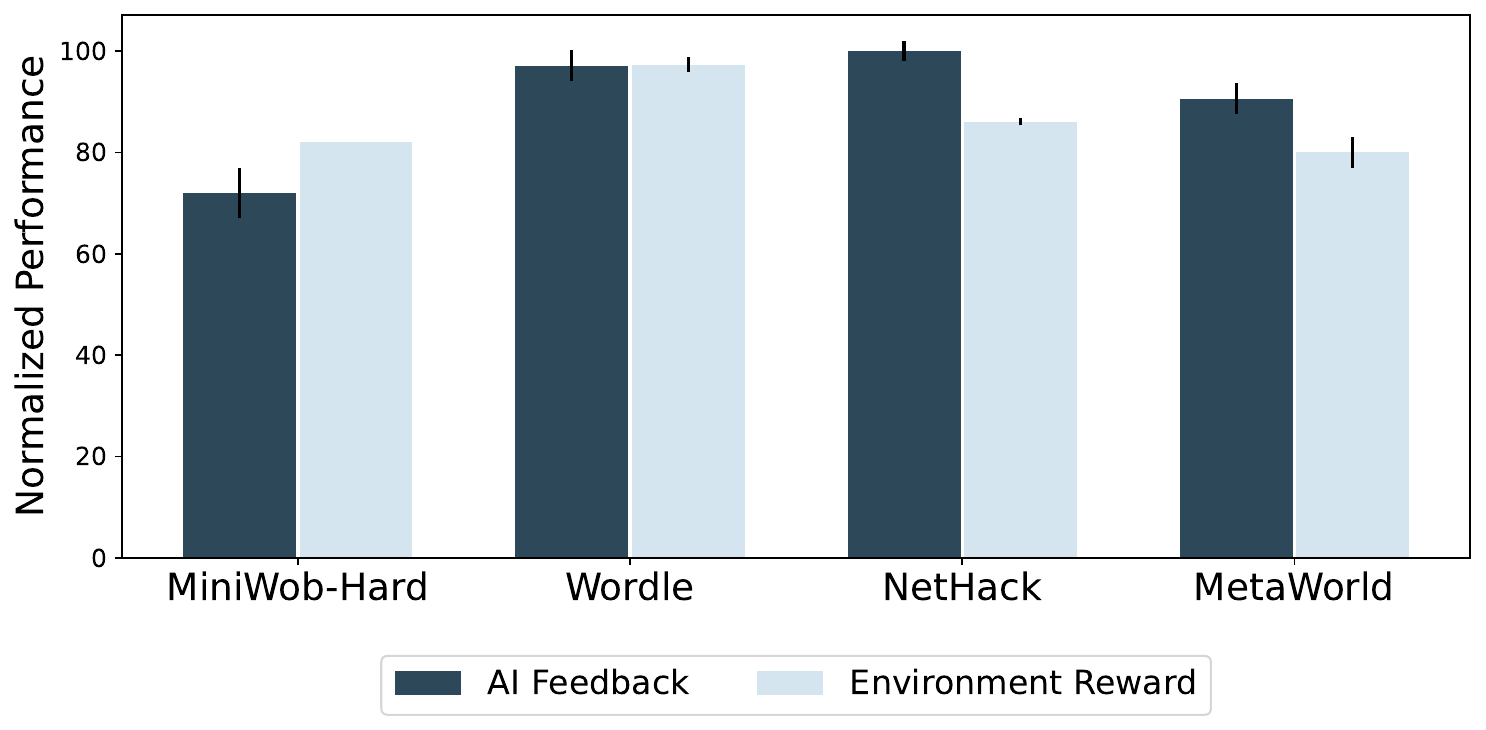}
    \caption{Comparison between the best performing LLM-based reward (AI Feedback) and human designed  rewards for each domain.}
    \label{fig:extrinsic}
\end{figure}

In Figure \ref{fig:extrinsic}, we compare the performance of an RL agent trained using a reward function derived from AI feedback with that of an agent trained on human-designed rewards across different environments. We observe that AI feedback achieves comparable results, with an average score of $89.93$ versus $86.3$ for the human-designed reward. The objective of this experiment is not to argue that LLM-based rewards consistently outperform human-crafted ones—since expert human knowledge can always be encoded into a reward function—but rather to contextualize the performance of LLM-based rewards. Notice that for MetaWorld we report the performance after fine-tuning the LLM as described in Section \ref{sec:finetuning}.

\subsection{AI feedback and heuristic functions}
\label{sec:heuristic}
While prior works have shown that rewards can be extracted from a \lm~\citep{brooks2024large, klissarov2024motif}, it can be more generally thought of as encoding a heuristic function $h$. The function $h$ contains high-level, multi-step information about the MDP $M$. To extract it, one can solve the re-shaped MDP $\tilde{M}$ with $\tilde{r}(s_t,a_t)=r(s_t,a_t)+(1-\lambda)\gamma \mathbb{E}_{s_{t+1}|s_t,a_t}[h(s_{t+1})]$ and $\tilde{\gamma} = \lambda \gamma$ where $\lambda \in [0, 1]$~\cite{cheng2021heuristic}. Solving $\tilde{M}$ yields a policy $\pi^*$ that is also optimal in $M$ - its value function's bias can be shown to converge to $V^*$ in $M$ as a function of $||h-V^*||_\infty$.

Specifically, assume access to an initial dataset $\cD_0$, from which a heuristic $h$ can be computed. In the reshaped MDP $\tilde{M}$, one can learn a new policy $\pi$ which optimizes $\tilde{r}$ with $\lambda \in [0, 1]$.~\Cref{eq:hurl_regret} shows the performance difference lemma~\cite{kakade2002approximately} as a function of true and reshaped MDP quantities:

\begin{equation}
\begin{split}
\cL(\pi, h)=&\mathbb{E}_{\cD_0}[V^*(s)-V^\pi(s)]\\
    =&c_1 \color{RoyalBlue}\mathbb{E}_{\cD_0}\bigg[\tilde{V}^*(s)-\tilde{V}^\pi(s)\bigg]\color{black}+c_2\color{ForestGreen}\mathbb{E}_{\cD^\pi}\bigg[\tilde{V}^*(s)-\tilde{V}^\pi(s)\bigg]\color{black}+c_3\color{WildStrawberry}\mathbb{E}_{\cD^\pi}\bigg[h(s')-\tilde{V}^*(s')\bigg]\color{black},
\end{split}
\label{eq:hurl_regret}
\end{equation}
where $c_1,c_2,c_3$ are non-negative constants. Minimizing $\cL(\pi, h)$ with respect to $\pi$ and $h$ can be achieved by minimizing each individual term. In particular, the red term suggests that the heuristic $h$ has to be updated on data from $\cD^\pi$ in order to not become "stale". This points out a shortcoming of existing LLM-as-critic algorithms, which sometimes fix $h$ after distilling the \lm knowledge into it~\cite{klissarov2024motif}

These theoretical findings suggest, in particular, that heuristic $h$ (in our case, the Bradley-Terry preference model), has to be updated with on-policy samples, similarly to empirical results from~\Cref{fig:exploration_nethack}.

\subsection{Additional Considerations for Preference-based Reward Modeling}
\label{sec:nonmarkovian}

\begin{figure}[ht!]
    \centering
    \includegraphics[width=0.85\linewidth]{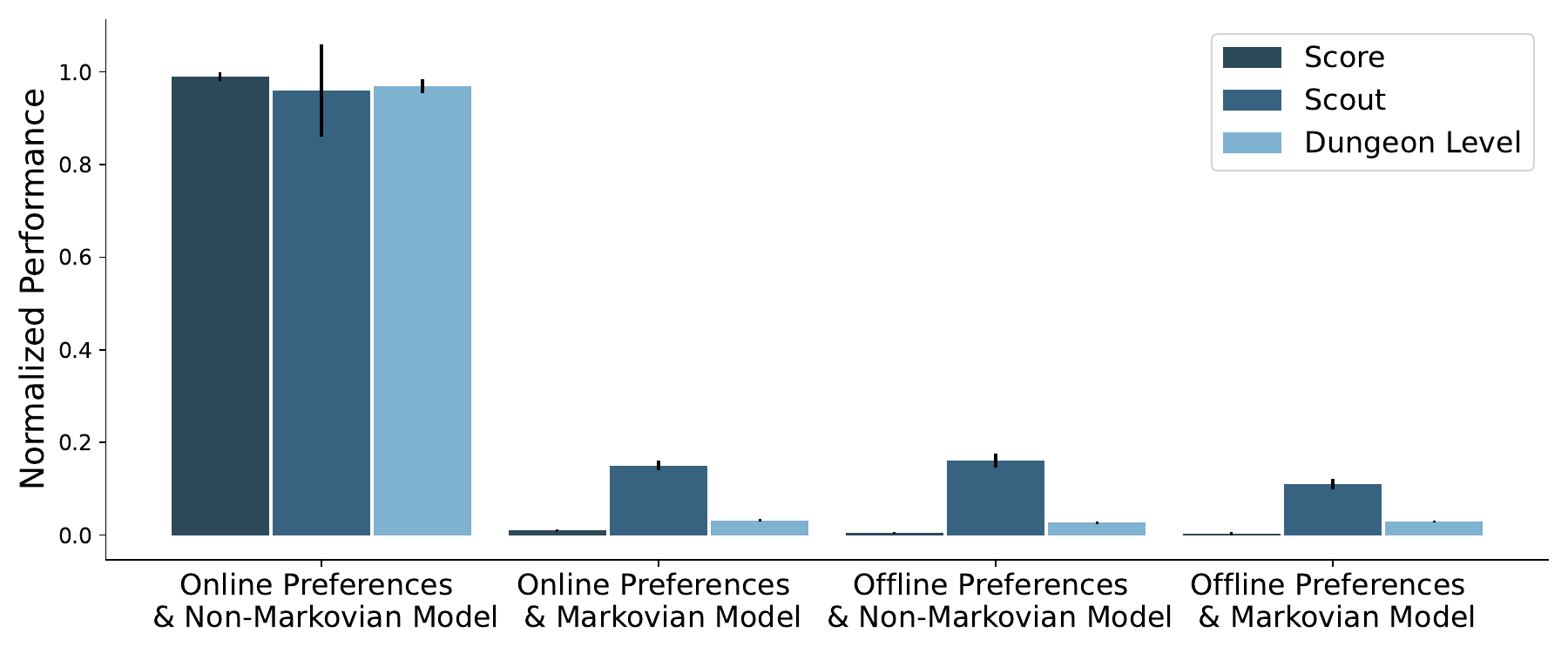}
    \caption{Successful exploration on Nethack depends on both online preference elicitation and a non-Markovian reward function.}
    \label{fig:additional_nethack}
\end{figure}

In Figure \ref{fig:additional_nethack}, we present the properties that were important to obtain effective exploration on NetHack, without the counting term shown in Equation \ref{eq:rint}.

\subsection{In-Context Learning for Reward Modeling}
\label{sec:incontext_reward}
\begin{figure}[ht!]
\centering
    \begin{subfigure}[t]{0.45\linewidth}
        \centering
        \includegraphics[width=\linewidth]{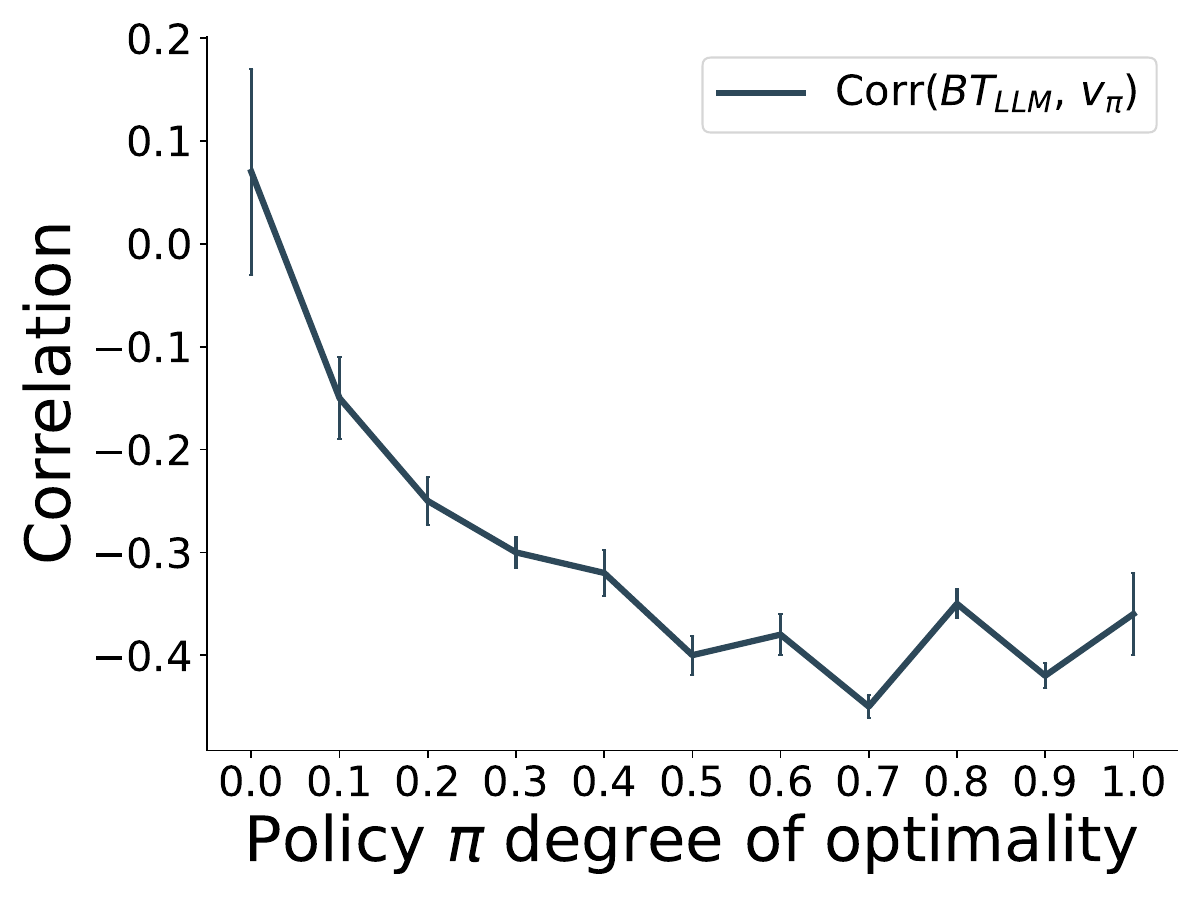}
        \caption{Before in-context learning}
        \label{fig:before_incont}
    \end{subfigure}
    \begin{subfigure}[t]{0.45\linewidth}
        \centering
        \includegraphics[width=\linewidth]{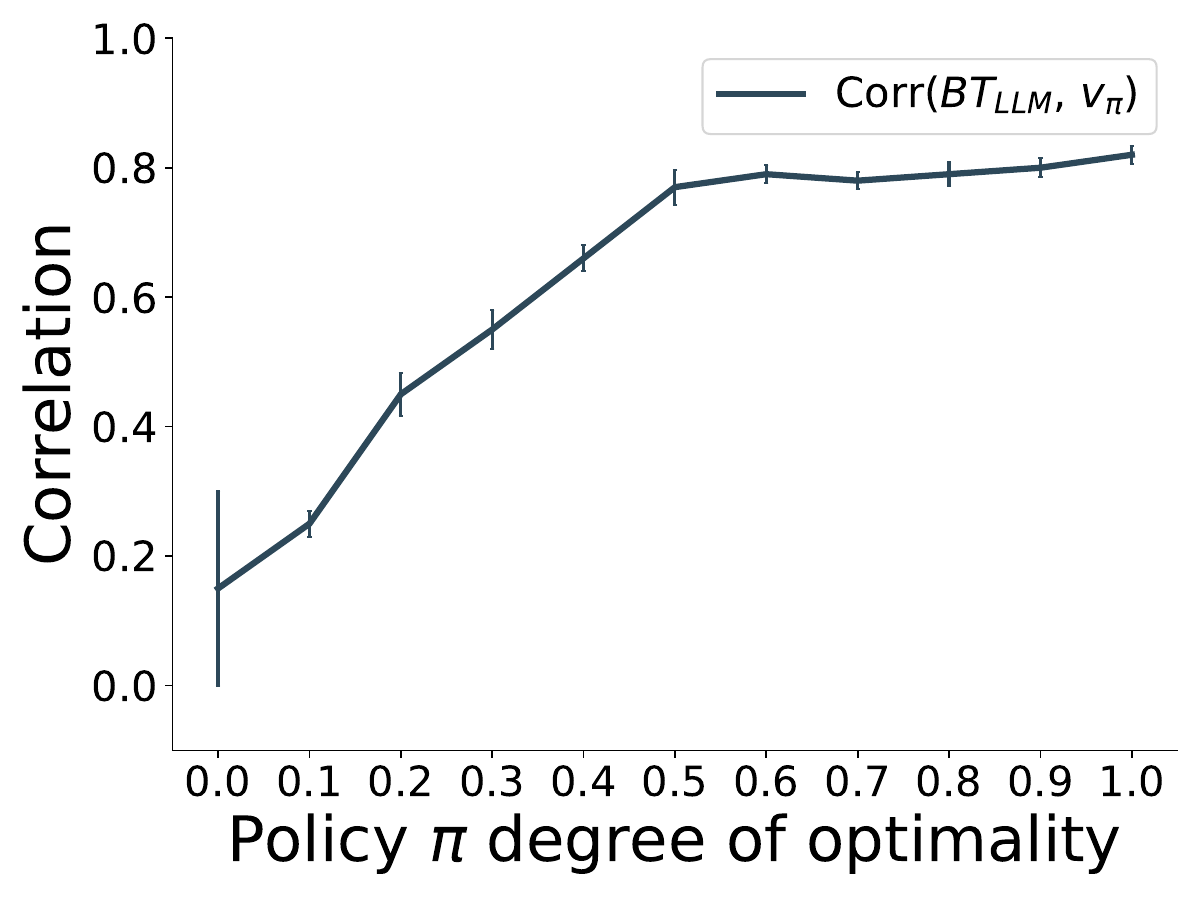}
        \caption{After in-context learning}
        \label{fig:incont}
    \end{subfigure}
    \caption{\textbf{AI feedback can be adapted to novel settings through in-context learning.} While the original LLM does poorly on Eldrow due to out-of-distribution, it manages to correct its feedback the task using in-context hints.}
    \label{fig:eldrow}
\end{figure}

In Figure \ref{fig:eldrow}, we present a variation on the Wordle game where the color code has been altered, which we refer to as Eldrow (reverse Wordle). Under this transformation, the off-the-shelf model provides feedback that correlates very poorly with the optimal value function. When we measure the perplexity of the LLM on a natural language description of the new rule set of Eldrow (see Appendix \ref{sec:app}) we obtain a value of $6.97$ which is higher than the one measured on the standard rule set of Wordle, with a value of $5.06$. Given that the difference in values is not very large, we leverage the simplest way for adapting the LLM: through in-context learning. As shown in Figure \ref{fig:incont}, by providing hints in the prompt about the new rule set, the LLM adapts its preferences and generates a Bradley-Terry model that recovers the correlation values we witnessed in \ref{fig:corr_vf}.

\subsection{LLMs as novelty detectors}
We hypothesize that LLMs with long contexts can effectively act as novelty detectors. Within the scope of RL problems, this implies the ability to tell, for example, whether a sub-trajectory is contained in the replay buffer.

\begin{figure}[h!]
    \centering
    \includegraphics[width=0.84\linewidth]{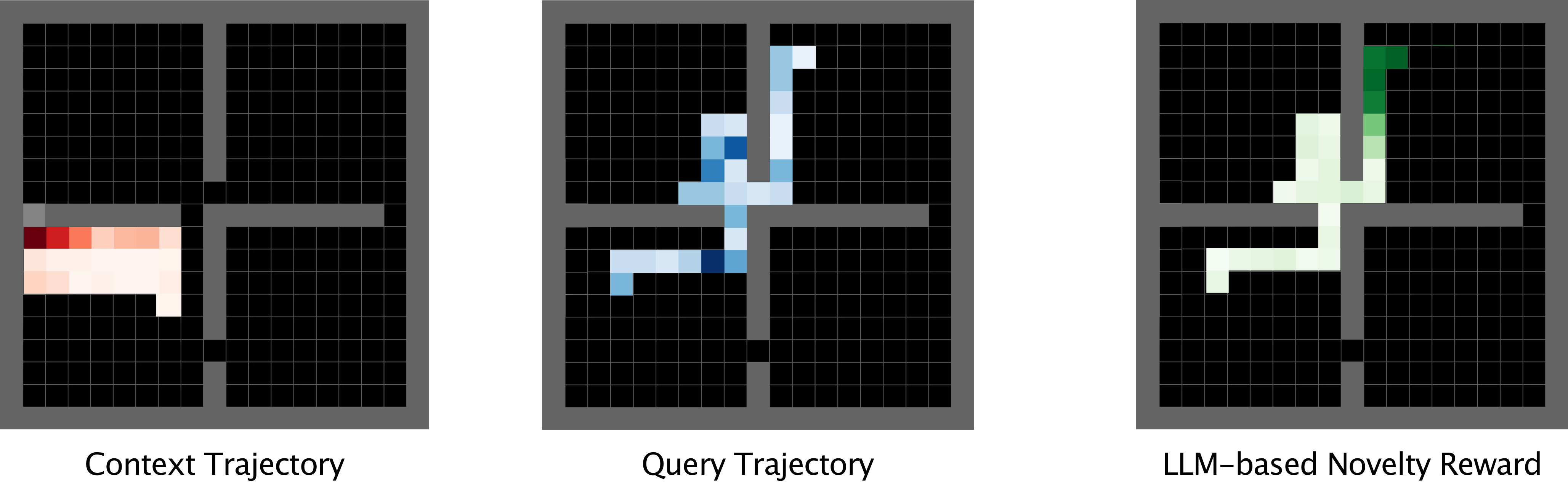}
    \caption{\textbf{LLMs can capture observation novelty.} Given the context trajectory (red), and a single observation sampled uniformly at random from the query trajectory (blue), the LLM correctly identifies novel states that are seen in the query but not in the context (green).}
    \label{fig:fourrooms_exploration}
\end{figure}

To test this, we query \texttt{Gemini-1.5 Pro}~\citep{team2023gemini} with a context video containing 500 frames of an agent exploring the bottom-left room (\Cref{fig:fourrooms_exploration}-left) and a single frame sampled uniformly at random from a query episode which covers in the top-right room, center and bottom of the maze (\Cref{fig:fourrooms_exploration}-middle). We ask the LLM to identify novel query states, i.e. states which are not seen in the context episode. We then train a direct predictor (3-layer MLP) to estimate the probability of any state on the grid to be novel with respect to the context (\Cref{fig:fourrooms_exploration}-right). The \lm correctly identifies the top-right portion of the trajectory to be novel, knowledge which could then be used to construct an intrinsic reward function.


%% file: sections/4-Related_work.tex
Large language models (LLMs) require additional adaptation for general-use language tasks~\citep{christiano2017deep, stiennon2020learning, ouyang2022training, mialon2023augmentedlanguagemodelssurvey}. 
Without additional context and/or fine-tuning, LLMs can generate misleading, harmful, or even nonsensical answers to queries or conversations with humans~\citep{bai2022constitutional}.
To modify their behavior, it is necessary to tune their prompts and/or fine-tune their outputs to ensure their output is desirable w.r.t. some set of linguistic tasks before deployment.
This at least if not more true in embodied settings, where real-world actions can have physical consequences, and methodologies for modifying LLM behavior in embodied settings more-or-less align with efforts in the language space.

\paragraph{Prompt tuning} Arguably the most common theme among techniques that modify LLM behavior in general is to change the prompt such that the distribution of LLM outputs better-fits a given desiderata on behavior. 
Prompt-engineering can greatly align or calibrate an LLM, pretrained or no, to desired beneficial behavior~\citep{christiano2017deep, glaese2022improving, bai2022constitutional}, or even expose harmful or other unexpected behaviors.
Chain-of-thought~\citep[CoT, ][]{Wei2022ChainOT} is an in-context method to either few-shot or zero-shot~\citep{kojima2022large} adjust an LLM's outputs to generate more correct responses to question-and-answering tasks.
Further modifications to the prompt such as providing feedback from an environment~\citep{yao2022react}, self-critique~\citep{zelikman2022star}, or self-reflection~\citep{shinn2023reflexion} can improve LLM performance in language as well as tasks that have an environment.
The biggest promise of in-context-based methods in RL is that somewhere within the \llm's conditional distribution is the optimal policy for any given task~\citep{brohan2023rt2visionlanguageactionmodelstransfer, szot2023large}, an accurate world-explicit model~\citep{lin2024learningmodelworldlanguage}, and/or a useful reward-model~\citep{klissarov2024motif}.
However, it is at best speculative as LLM's are black box systems and prompt optimization is extremely difficult, and besides: systems built on this idea still must still overcome affordance mismatch~\citep{Ahn2022DoAI} and hallucinations~\citep{llm_hallu_snowball} to be useful for RL.


\textbf{Querying model for feedback} Another hypothesis is that LLMs contain knowledge relevant to tasks, and this knowledge can be extracted~\citep{xu2024largelanguagemodelsgenerative} in a way to train a policy that has desirable behavior~\citep{huang2022language}.
RL AI Feedback~\citep[RLAIF][]{bai2022constitutional, lee2023rlaif} is a scalable method akin to but without the practical issues  that come paired with RL from Human Feedback~\citep[RLHF][]{christiano2017deep}, the goal of which is to fine-tune an existing LLM to be more specific, accurate, innocuous, etc.
RLAIF trains a reward model on a dataset collected from an LLM's preferences given a dataset of language responses from an LLM and a given set of queries, and this reward model is used to train a policy using RL, for example using PPO.
This process of extracting knowledge using preference data can also be directly used to train a policy without a reward model~\citep{rafailov2024directpreferenceoptimizationlanguage}. 



